\documentclass[letterpaper, 10 pt, conference]{ieeeconf}  

\IEEEoverridecommandlockouts                              

\usepackage{times}
\usepackage{soul}
\usepackage{url}
\usepackage[utf8]{inputenc}
\usepackage[hidelinks]{hyperref}
\usepackage{graphicx}
\usepackage{amsmath}
\usepackage{amssymb}
\usepackage{booktabs}
\usepackage{multicol}
\usepackage{multirow}
\usepackage{algorithm}
\usepackage{color}
\usepackage{algpseudocode}
\usepackage{xspace}
\usepackage[usenames,dvipsnames]{xcolor}
\usepackage[switch]{lineno}

\usepackage{subcaption}
\usepackage[export]{adjustbox}

\newcommand{\model}{{PhyPlan}\xspace}

\begin{document}

\title{\LARGE \bf
PhyPlan: Generalizable and Rapid Physical Task Planning with Physics-Informed Skill Networks for Robot Manipulators
}


\author{
Mudit Chopra$^{1*}$, 
Abhinav Barnawal$^{1*}$, 
Harshil Vagadia$^{2}$, 
Tamajit Banerjee$^{2}$,
Shreshth Tuli$^{2}$, \\
Souvik Chakraborty$^{1\#}$ and 
Rohan Paul$^{1\#}$ \\
\footnotesize $^{1}$Work primarily done when at IIT Delhi,
\footnotesize $^{2}$Affiliated with IIT Delhi,
\footnotesize $^{*}$Equal Contribution,
\footnotesize $^{\#}$Equal Advising.
}





\maketitle

\begin{abstract}
    Given the task of positioning a ball-like object to a goal region beyond direct reach, humans can often throw, slide, or rebound objects against the wall to attain the goal. However, enabling robots to reason similarly is non-trivial. Existing methods for physical reasoning are data-hungry and struggle with complexity and uncertainty inherent in the real world. This paper presents PhyPlan, a novel physics-informed planning framework that combines physics-informed neural networks (PINNs) with modified Monte Carlo Tree Search (MCTS) to enable embodied agents to perform dynamic physical tasks. PhyPlan leverages PINNs to simulate and predict outcomes of actions in a fast and accurate manner and uses MCTS for planning. It dynamically determines whether to consult a PINN-based simulator (coarse but fast) or engage directly with the actual environment (fine but slow) to determine optimal policy. Given an unseen task, PhyPlan can infer the sequence of actions and learn the latent parameters, resulting in a generalizable approach that can rapidly learn to perform novel physical tasks. Evaluation with robots in simulated 3D environments demonstrates the ability of our approach to solve 3D-physical reasoning tasks involving the composition of dynamic skills. Quantitatively, PhyPlan excels in several aspects: (i) it achieves lower regret when learning novel tasks compared to the state-of-the-art, (ii) it expedites skill learning and enhances the speed of physical reasoning, (iii) it demonstrates higher data efficiency compared to a physics un-informed approach. Videos of robot performing tasks are available at: \textcolor{blue}{\href{https://phyplan.github.io}{https://phyplan.github.io}}

\end{abstract}

\section{Introduction}
\label{sec:introduction}
Consider the scenario where the robot is asked to put a ball inside an empty box. The robot must improvise and dynamically interact with the objects in the environment if the box is directly unreachable. It may need to throw the ball, slide it across, or bounce it off a wedge to make it reach the box. Reasoning about the effects of various actions is a challenging task. Although the solution may appear self-evident to humans, a robot requires several demonstrations of each alternative to attain proficiency in a given task. Learning to reason with physical skills is a hallmark of intelligence. Increasing evidence is emerging that humans possess an intuitive physics engine involved in perceptual and goal-directed reasoning \cite{allen2020rapid,bear2021physion}. In essence, such reasoning requires (implicit or explicit) learning of predictive models of skills and the ability to chain them to attain a stated goal.


Recent efforts such as \cite{allen2020rapid} learn to solve physical tasks by exploring interactions in a physics simulator. The reliance on a realistic physical simulator during training slows the learning process. ~\cite{bakhtin2019phyre} benchmarks physical reasoning tasks, proposes alternative model-free approaches to solve such tasks and shows their limitation in long-horizon reasoning, mainly when rewards are sparse (using a physical tool in a specific way to acquire large rewards). This motivates an investigation into the availability and use of “physical priors” to scale skill learning to longer horizons. \cite{toussaint2018differentiable} addresses scalability by adopting symbolic abstractions over physical skills expressed as governing equations with free parameters learnable from data. Such symbolic abstractions can be embedded into a planning domain for PDDL-style planning to arrive at multi-step plans. However, modelling errors during hand-encoding of \emph{exact} equations leads to brittleness.

\begin{figure*}
    \centering
    \includegraphics[width=\linewidth]{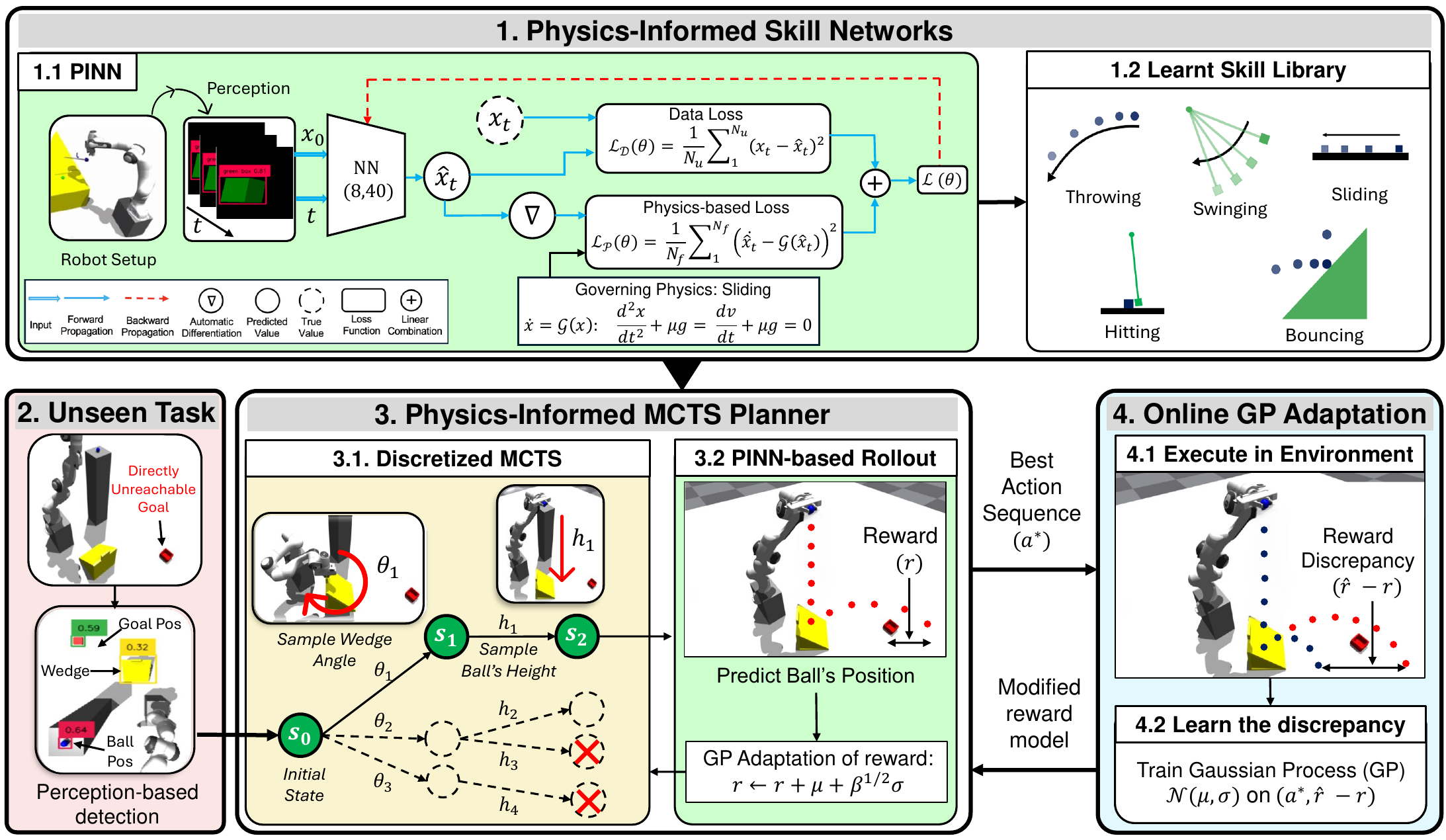}
    \caption{\footnotesize {\textbf{Approach Overview.} Clockwise from top. (1) The robot learns a model of physical skills such as hitting, swinging, sliding etc. using a Physics-Informed Neural Network (PINN). The network incorporates a coarse physics governing equation in its loss function and can predict the future trajectory of the interacting object as well as the latent physical parameters of the domain. Given a new task (2), a Monte-Carlo Tree Search (MCTS) (3) searches for a plan by exploring the space of skill compositions, i.e., sampling over continuous parameters e.g., the height of object release, the angle of a wedge with which the object may collide. Simulating the trajectories during physical interaction using a forward pass of a PINN is faster than simulating rich inter-object dynamic interactions in a high-fidelity physics simulator (with complex physical models). The MCTS plan search (4) periodically balances sampling with the PINNs with occasional rollouts in the high-fidelity simulator. The discrepancy in the reward between (i) predicted by PINN rollouts and rollout in the physics simulator is modelled using a Gaussian Process ($\mathcal{GP}$) that guides the MCTS towards the correct plan.}}
    \label{fig:motivation}
\end{figure*}

In response, this study introduces a physics-informed skill acquisition and planning model that efficiently extends to multi-step reasoning tasks, offering advantages in (i) data efficiency, (ii) rapid generalizable physical reasoning, and (iii) reduced training time compared to methods relying on intricate physics simulations. Simulation time exhibits exponential growth when chaining skills, particularly in continuous state spaces. To address this, we leverage a limited number of physical demonstrations within the simulation engine to acquire \textit{primitive} skills such as throwing, sliding, and rebounding from observed trajectories. Subsequently, we train a PINN-based predictive model. The acquired skills are then incorporated into a Monte Carlo Tree Search (MCTS) algorithm, facilitating the resolution of complex tasks through neural network simulations with minimal reliance on real-world simulations for fine-tuning (see Figure \ref{fig:motivation}). Our modification of the MCTS algorithm \ref{alg:mcts} chooses to either enquire about the PINN-based coarse simulator or the fine simulator and (ii) accounts for the inherent sparsity of rewards commonly encountered in tasks involving physical reasoning. We call this approach \model. We designed simulated 3D physical reasoning tasks involving the Franka Emika arm using the Isaac Gym tool~\cite{nvidia2022isaacgym} to evaluate our approach. These tasks involve physical skills like sliding, throwing, and swinging. Each task requires the robot to get a ball/puck to reach the goal region by performing actions. We evaluate the agent in unseen scenarios based on how close it got the ball/puck to the goal. Empirical evaluation demonstrates that our proposed approach outperforms conventional methods, such as Deep Q-Networks (DQN), while requiring a notably reduced number of training simulations.

\section{Related Works}
\label{sec:related-works}
Within robotics, several efforts have focused on learning physical skills using reinforcement learning or learning from demonstration.~\cite{park2020inferring} use inverse RL to learn object insertion trajectories.~\cite{zeng2020tossingbot} learn visuo-motor policies for throwing arbitrary objects.~\cite{chen2022towards} focus on object transfer using catching. Instead of learning a predictive model for specific skills, others build on Dynamic Movement Primitives~\cite{ijspeert2013dynamical} as a generic skill representation to guide policy learning. Although effective for learning smooth~\cite{bahl2020neural} or periodic dynamics~\cite{yang2022learning}, the representation is less expressive for modelling contact or collision interactions. Although successful in learning short-horizon skills, scalability to long-horizon tasks is limited.\\
\indent Another body of work focuses on learning to compose skills for attaining goal-directed tasks. Closely related is the work of~\cite{toussaint2018differentiable}, where a planning framework is introduced for sequential manipulation using physical interactions such as hitting and throwing. Wherein this approach relies on exact physics equations, our approach learns a predictive model for skills using coarse physics and data, allowing more resilience to noise and accelerated training. Further, embedding skills into a model-based reinforcement learning approach facilitates solving multi-step tasks and adapting online to the environment's unmodeled aspects. The work of~\cite{zeng2020tossingbot} introduces a trial and error method predicting residuals on top of control parameters specifying a skill parameterization. \cite{moses2020visual} proposed to learn the latent reward function modelled as a Gaussian Process ($\mathcal{GP}$) and trained using the rewards observed during training. The learnt function is a prior where corrections are learnt when the robot interacts in the real environment, an approach we adopt in this paper.
A concurrent development bridging the realms of physics and data science is the Physics-Informed Neural Network (PINN)~\cite{raissi2019physics}; the idea here is to impose regularisation on the latent space to ensure adherence to the governing physics. PINN offers distinct advantages, primarily driven by (i) its computational efficiency, thanks to the final model's neural network structure, and (ii) its remarkable generalisation capabilities derived from incorporating physics as a regularisation factor. Improvements over the vanilla PINN includes multi-fidelity PINN~\cite{chakraborty2021transfer}, gradient-enhanced PINN~\cite{yu2022gradient}, and gradient-free PINN~\cite{navaneeth2023stochastic}. In this work, we harness the power of PINN to model the foundational `skills' necessary for executing complex physical reasoning `tasks'.\\
\indent Recent works have explored the Large Language Models (LLMs) as planners. While \cite{gao2023physically} uses LLMs to generate high-level task plans, \cite{hao2023reasoning} augments the LLM with a world model and reasons with MCTS to produce high-reward reasoning traces. These works operate on discrete action spaces. Reasoning in continuous domain is still challenging for LLMs mainly because of 
the incapability to balance exploration vs. exploitation to efficiently explore vast reasoning space \cite{hao2023reasoning}.

\section{Problem Setup}
\label{sec:problem-formulation}
We consider a robot manipulator operating in a tabletop environment. Assume that the robot can position its end-effector at a desired configuration, grasp an object, move it while holding it, and release the object-oriented at a target pose. The environment consists of ball-like object (puck/ball) and additional objects such as a wedge, a flat surface (such as a table), a positioning plank (for instance, a bridge), a box-like object (goal), a pendulum-like assembly capable of swinging a pendulum that rotates on a hinge. The ball-like object can interact with other objects, e.g., by sliding, colliding, or falling under gravity, finally coming to rest on a ground surface or inside a container-like object.
The robot's objective is to ensure that the dynamic object reaches a goal region $g$. Note that direct grasping and releasing the ball-like object over the goal may not be feasible, particularly when the goal is far from the robot. Hence, the robot may interact with additional objects as tools, \textit{i.e.}, position and release points in continuous space, such that when the ball-like object is released, the dynamic interactions facilitate reaching goal region. For instance, a ball resting on a table may be hit by a pendulum, slide on a table, and fall into a goal region via projectile motion.

The action space in our setting is continuous and the rewards are sparse, making the problem inherently hard to generalise over novel task-settings. We, therefore, decompose the task at hand into components with known skills, predict the value of an action in the task by composing the skills and run MCTS over the prior. We periodically learn the errors in the prior as a $\mathcal{GP}$.
Formally, let $s_t \in \mathcal{S}$ denote the world (a.k.a. environment) state at time $t$, observed as an image $i_t$ from a depth camera. Let $\mathcal{O}$ denote the set of objects in the environment, where each $o \in \mathcal{O}$ has a set of continuous modes, $M(o)$ (degrees of freedom) for interacting with the environment. For example, a wedge may be placed at a coordinate $(x, y)$ and oriented at angle $q$, leading to $M(wedge) = \{x, y, q\}$. Similarly, the interaction modes for a pendulum can be plane angle and release angle. For a ball, it may be velocity $v$ and coordinates of release $h$. We represent the action (equiv. action sequence) taken on state $s_t$ as $A_t \in \mathcal{A}$. Action outcomes are modelled as stochastic transition model $s_{t+1} \sim T(s_t, a_t)$. The goal regions are embedded in the reward model
$R: \mathcal{S} \times \mathcal{A} \times \mathcal{S} \mapsto \mathbb{R}$.


We learn a goal-conditioned policy $A_t \sim \pi_g(s_t)$, where $A_t$ is a tuple of sub-actions $(a_0, a_2, \ldots, a_N)$ indicating the setting of interactive variables for each object, such as $A_t = \{M(o_1), M(o_2), \ldots, M(o_n)\}$. The robot performs action $A_t$ only based on $s_t$, obtains the reward, updates the reward model, and resets the state. In essence, the robot sets the configurations of the necessary objects to the desired values and releases the ball-like object. This is considered as a single ``trial''. We do not consider future states like $s_{t+1}$. Hence, our setup is one of Contextual Multi-armed Bandits setup (C-MAB)~\cite{lu2010contextual}. Our policy aims to minimise the regret $e = \tfrac{r - r*}{r}$, where $r$ is the maximum reward obtained by the agent in a trial, and $r*$ is the maximum attainable reward. 



\section{Technical Approach}
\label{sec:technical-approach}
Our approach is focused on factoring the generalized problem of physical reasoning as one of learning a skill model (physics-informed skill networks) that determines how an object engages with the environment and then using this model to plan multi-step dynamic interactions to reach the intended goal (Figure \ref{fig:motivation}). We postulate that learnt skill models would provide a strong prior for the planner and help make informed decisions. 

\subsection{Physics-Informed Skill Networks}
We consider the problem of learning a model for physical skills such as \emph{bouncing} a ball-like object off a wedge, \emph{sliding} over a surface, \emph{swinging} a pendulum, \emph{throwing} an object as a projectile and \emph{hitting} an object with a pendulum. Here, we interpret a physical skill as a model that predicts the state trajectory of an object as it undergoes a dynamic interaction with another object. Consider an object having state vector $x \in \mathbb{R}^n$ comprising of relevant physical quantities (e.g., position, velocity) performing a skill. Guiding physics of each skill can be captured by representing the governing differential equation in state space method $\dot{x} = \mathcal{G}(x)$, where $\dot{x}$ represents the time derivative of the state vector $x$. For instance, skills like throwing can be modelled as projectile motion with set of governing equations: $d^2y/dt^2 + g = 0$ (motion in vertical axis); $dx/dt - vx_{init} = 0$ (motion in horizontal plane). Here, $x$ and $y$ represent instantaneous position w.r.t. inertial frame in the horizontal and vertical direction, respectively, and $vx_{init}$ represents the initial horizontal velocity. Similarly, swinging can be modelled by pendulum dynamics and sliding by friction dynamics.

\begin{figure}[t]
    \centering
    {\includegraphics[width=\linewidth]{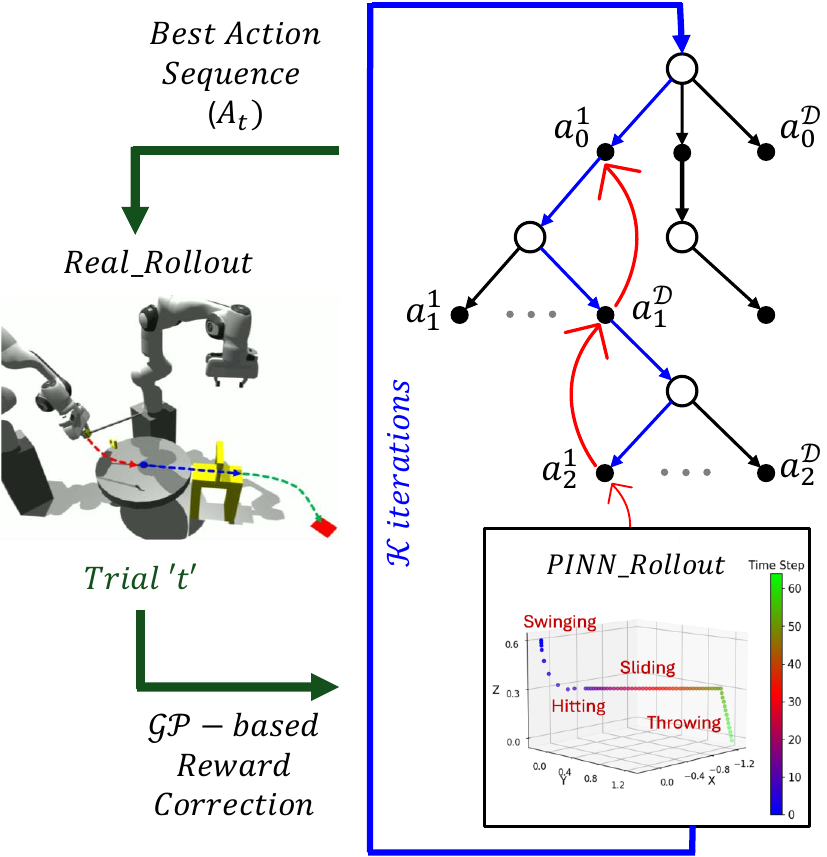}}
    \caption{\footnotesize \textbf{Adaptive Physical Task Planning in \emph{Bridge} task.} Let sub-actions $a_0, a_1, a_2$ represent the continuous space of bridge's orientation, pendulum's plane, and release angle, respectively. PINN-MCTS tree selects sub-action $a^j_i$ at depth $i$; predicts the value of the selected action sequence using \emph{PINN\_Rollout}; updates individual sub-action values and repeats for $\mathcal{K}$ iterations. Then, the best action sequence $A_t$ is executed in the environment, and reward model is updated using Gaussian-Process ($\mathcal{GP}$) for $t \leq T$ trials.
    }
    \label{fig:planner_diag}
\end{figure}

Various numerical integration schemes exist which can be employed to solve for governing differential equations. However, these techniques are often computationally expensive and time-consuming. Hence, we learn a function $\mathcal{F}: x_{o},t \rightarrow {x}_{t}$ mapping (object's initial state $x_{o}$, time $t$) to (object's state ${x}_{t}$ at time $t$). We model $\mathcal{F}$ as a Physics-Informed Neural Network $\mathcal{F_{\theta}}(x_{o}, t)$ with trainable parameters $\theta$ (Figure \ref{fig:motivation}).
For rapid skill learning from a sparse training dataset with $N_u$ data points, we constrain the latent space by incorporating governing equation $\dot{x} = \mathcal{G}(x)$ as a physics-informed loss term $\mathcal{L_P}(\theta)$. To compute $\mathcal{L_P}(\theta)$, we sample $N_p >> N_u$ points in the latent space, called collocation points, and use the equation
\begin{equation}
    \mathcal{L_P}(\theta) = \frac{1}{N_p} \sum_{i=1}^{N_p} (\dot{\hat{x}}_{t}^{i} - \mathcal{G}(\hat{x}_{t}^{i}))
\end{equation}
We also employ a data-driven MSE loss i.e.
\begin{equation}
    \mathcal{L_D}(\theta) = \frac{1}{N_u} \sum _{i=1}^{N_u} (x_{t}^{i} -\hat{x}_{t}^{i})^2
    \label{eq:data}
\end{equation}
Hence the net loss function is a linear combination of the physics-informed loss and the data-driven loss, i.e. 
$\mathcal{L}(\theta) = \mathcal{L_D}(\theta) + \epsilon \mathcal{L_P}(\theta)$, 
where $\epsilon$ is weighting factor obtained via grid search. Further, note that, we learn \emph{hitting} and \emph{bouncing} skills directly from data due of complex and intractable physics of their motion.

In practice, the parameters associated with physics models are not known apriori. For example, the coefficient of friction $\mu$ is often unknown. In such generalised settings, our function mapping can formulated as $\mathcal{F}: x_{o},t,\lambda \rightarrow {x}_{t}$ to account for variable parameters (represented by $\lambda$) as input. For skill learning in this setting we modify Physics Informed Neural Network as $\mathcal{F_{\theta}}(x_{o}, t,\lambda)$ so as to generalize over the physical parameters involved. Also, often one needs to predict these unknown variables parameters. This is known as inverse problem and can be efficiently solved by PINN. Consider the case of unknown coefficient of friction $\mu$. The inverse problem can be formulated as optimization problem: $\mu^{*}, \theta^{*} =  \underset{\mu,\theta}{argmin} (L(\mu,\theta))$. For determining the coefficient, PINN treats the unknown parameter as additional trainable parameters (absorbed in $\theta$) and estimates the same by solving the aforementioned optimisation problem. 

\subsection{Generalized Physical Task Planning}

Given an unseen physical reasoning task, we present a rapid and generalizable approach for multi-step reasoning to achieve the goal. Our approach consists of three procedures each described in Algorithm \ref{alg:mcts}. \textsc{Gen\_Plan} assumes the set of sub-actions $A$ (defined in section \ref{sec:problem-formulation}), such as bridge's orientation angle, pendulum's plane and release angle in \emph{Bridge} task, and runs trials to perform the task. \color{Brown} \texttt{Gen\_Action\_Samples($\mathcal{D}$)} \color{black} samples $\mathcal{D}$ (Discretization Factor, here $20$) values from the continuous space of each sub-action (\{$a^j_i\}^\mathcal{D}_{j=1}$ for $i^{th}$ sub-action) randomly from a uniform distribution. Procedure \textsc{Gen\_Action} \color{blue} \emph{reasons} \color{black} the best value among the samples for each sub-action using an MCTS planning algorithm (\textsc{PINN\_MCTS}).

\begin{figure*}[tb]
    \centering
    \begin{minipage}{0.24\linewidth}
         \includegraphics[width=\linewidth]{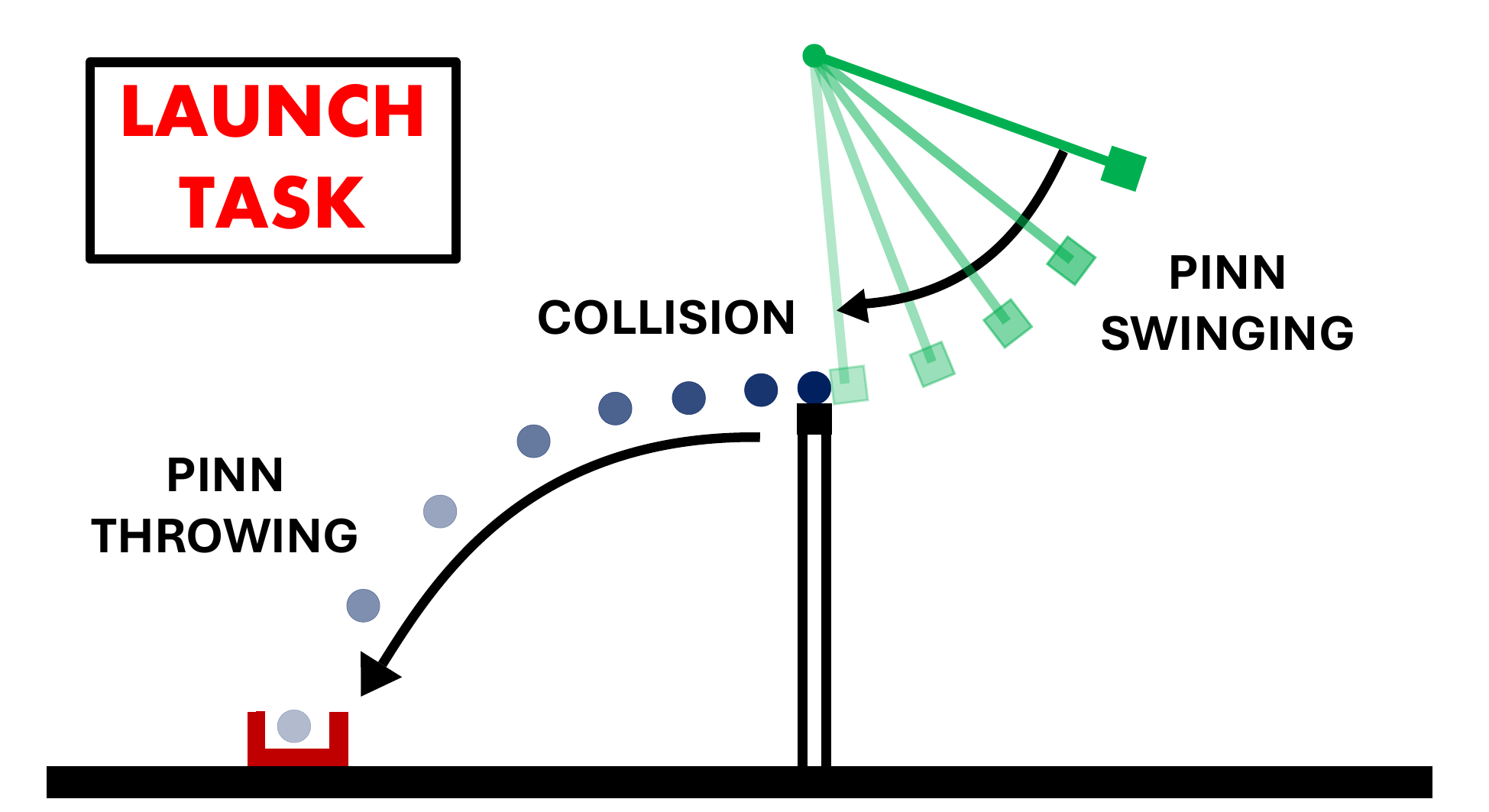}
    \end{minipage}\hfill
    \begin{minipage}{0.24\linewidth}
         \includegraphics[width=\linewidth]{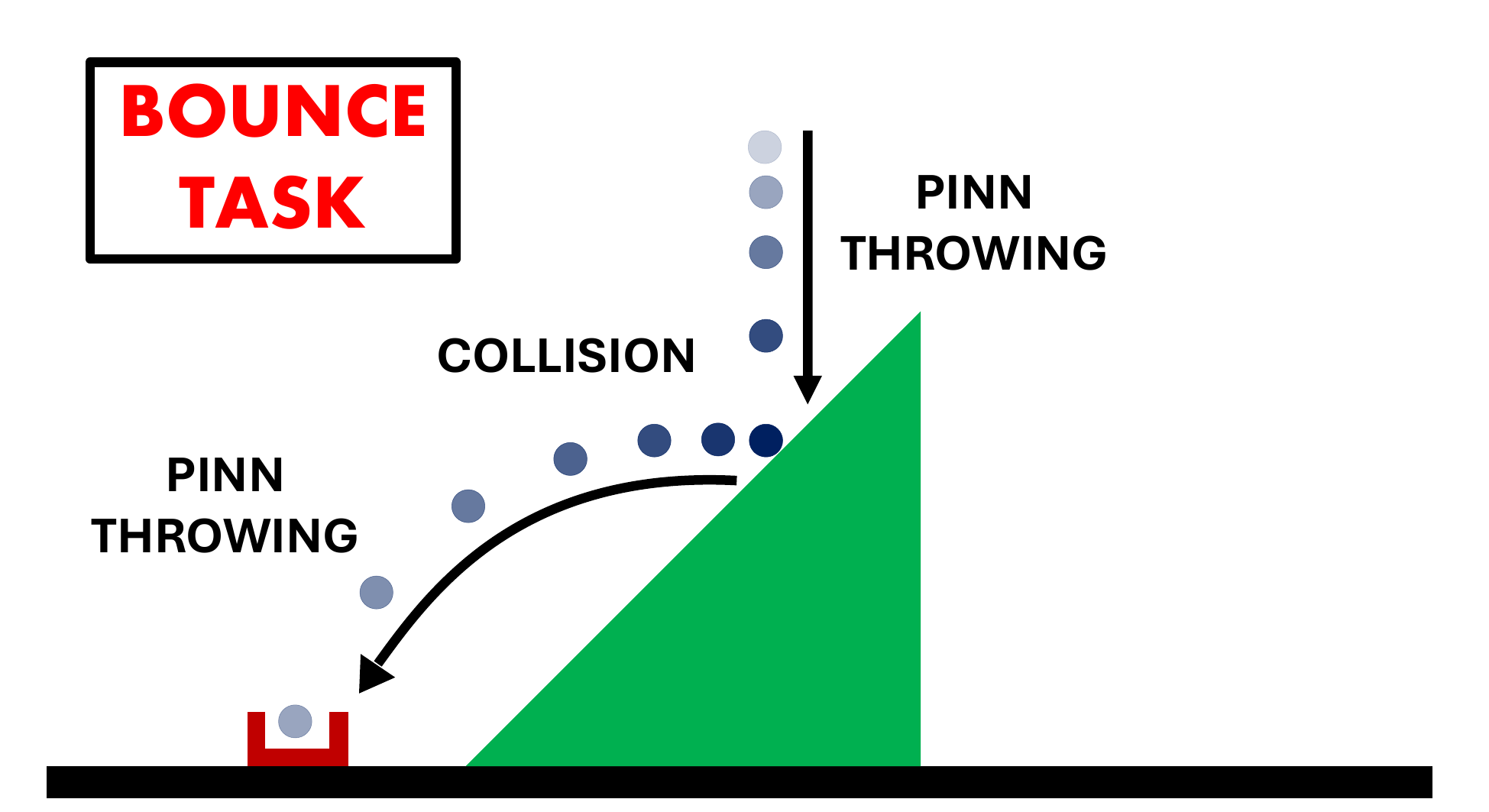} 
    \end{minipage}\hfill
    \begin{minipage}{0.24\linewidth}
         \includegraphics[width=\linewidth]{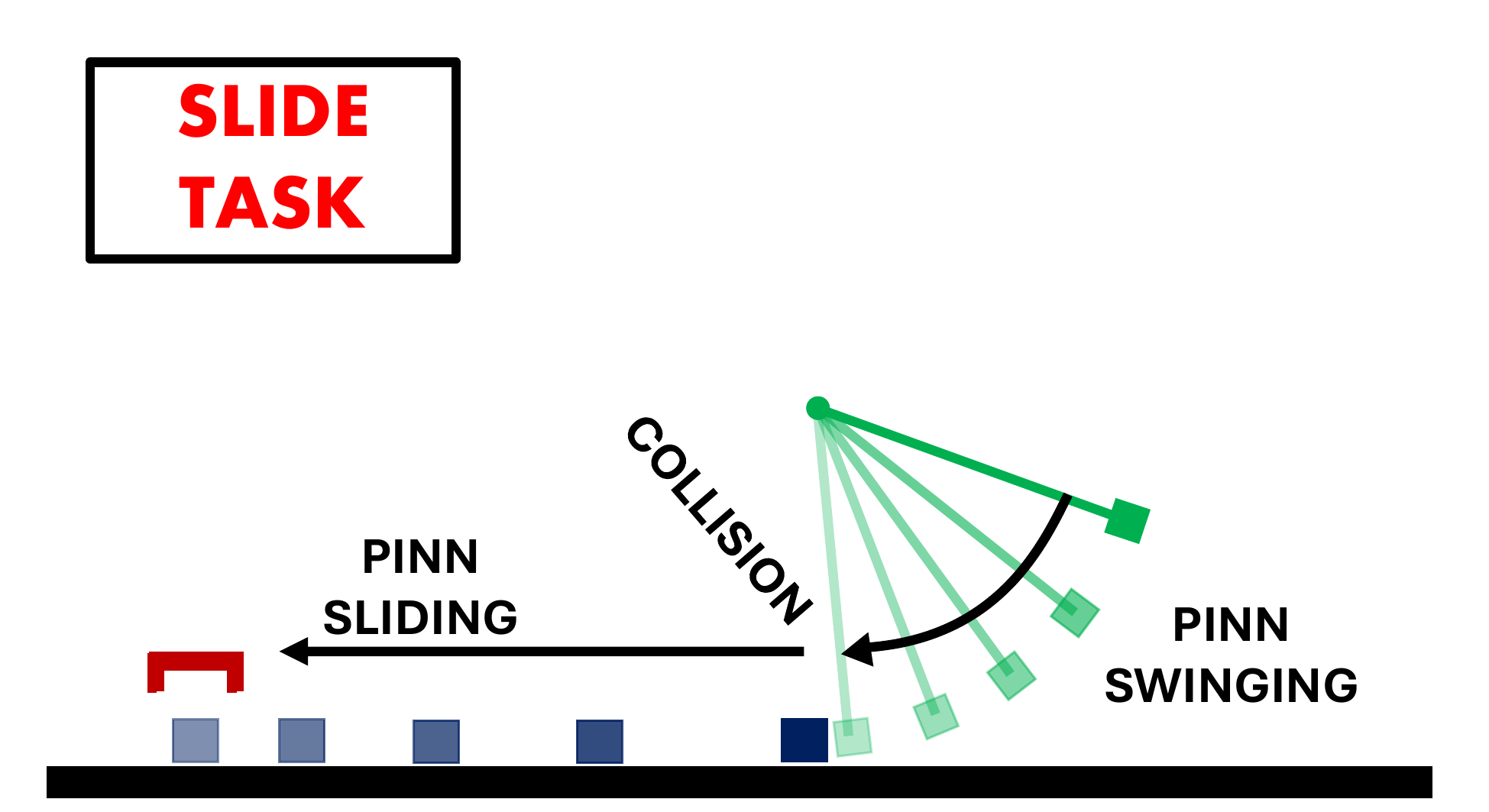} 
    \end{minipage}\hfill
    \begin{minipage}{0.24\linewidth}
         \includegraphics[width=\linewidth]{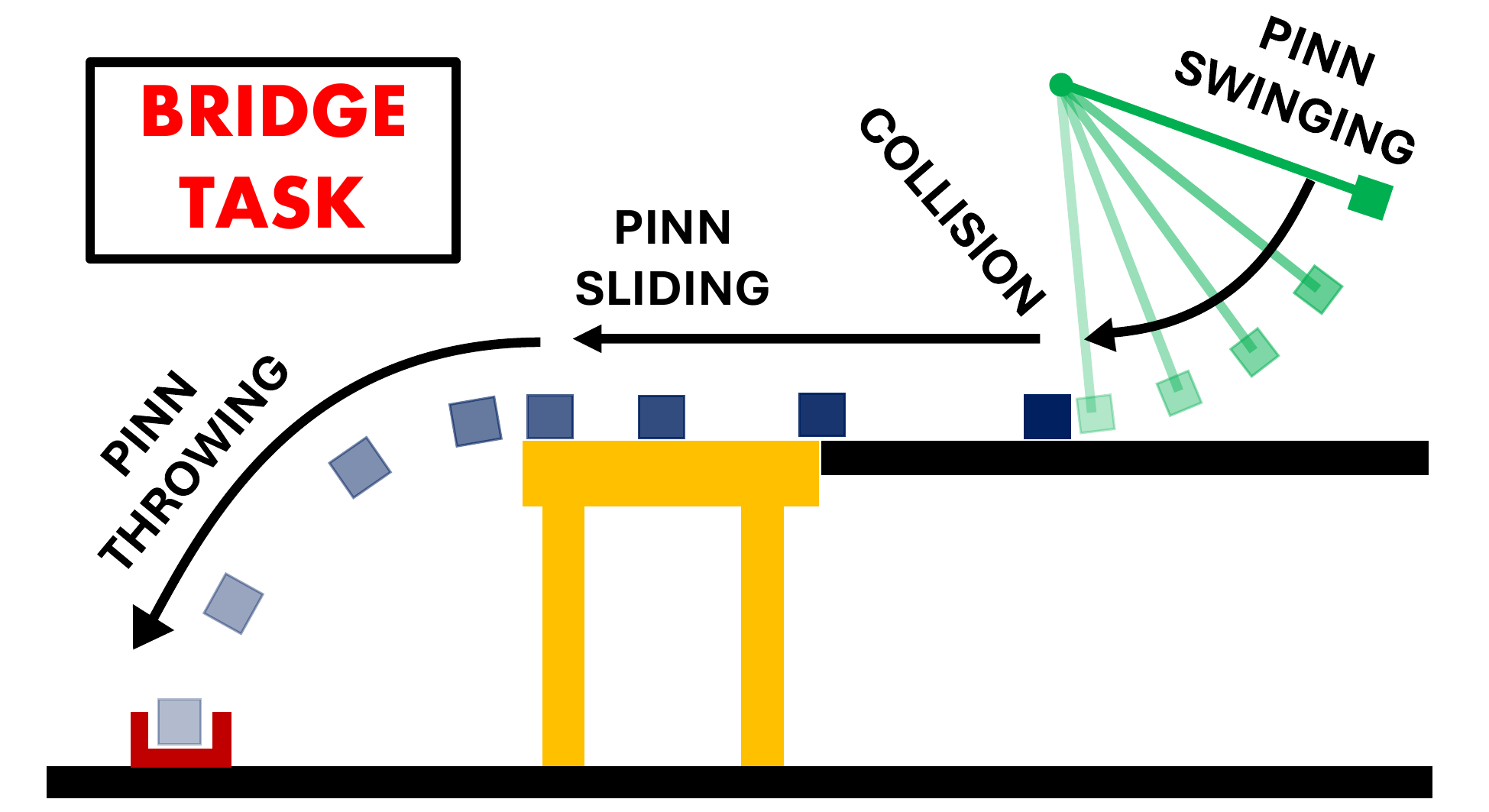}
    \end{minipage}
    \caption{\footnotesize \textbf{Benchmark Physical Reasoning Tasks.} This paper considers four tasks (shown above) inspired from ~\cite{allen2020rapid} and ~\cite{bakhtin2019phyre}. Each tasks requires the robot to place a dynamic object (a movable ball) be placed in a goal region (a container). The ball is not in direct kinematic range of the robot. Hence, the robot must make use of sequential dynamic interactions such as hitting, sliding, rebounding etc. so that the ball can land close to the goal. Further, physical parameters such as the coefficient of friction are latent and must be inferred allowing the robot to generalize to new settings.}
    \label{fig:skill-chaining} 
\end{figure*}

\begin{algorithm}[t]
    \small
    \caption{\textbf{: Adaptive Physical Task Planning}}
    \label{alg:mcts}
    \begin{algorithmic}[1] 
    \Procedure{Gen\_Plan}{}
        \State $x_g \leftarrow$ \texttt{Goal\_Pos()}
        \State $A = a_{0:N} \leftarrow$ \color{Brown} \texttt{Gen\_Action\_Samples}($\mathcal{D}$) \color{black}
        \For {$t = 1\ldots T$} \color{violet} \Comment{Trials} \color{black}
            \State $r^{sim}_t, A_t \leftarrow$ \textsc{Gen\_Action} $(A, x_g, \mu_{t-1}, \sigma_{t-1})$
            \State $r^{real}_t \leftarrow$ \color{blue} \texttt{Real\_Rollout}($A_t$) \color{black}
            \State $\eta_t \leftarrow r^{real}_t - r^{sim}_t$ \color{violet} \Comment{Reward Discrepancy} \color{black}
            \State $(\mu_t, \sigma_t) \leftarrow \mathcal{GP}(A_t, \eta_t;\ \mu_{t-1}, \sigma_{t-1})$ \color{violet} \Comment{Adapt} \color{black}
        \EndFor
    \EndProcedure\\
    
    \Procedure{Gen\_Action}{$a_{0:N},\ x_g,\ \mu,\ \sigma$}
        \State \color{teal} // $a_{0:N}$ :- sub-action sequence from task decomposition \color{black}
        \State \color{teal} // $x_g$ :- goal position; $\mu, \sigma$ :- mean, variance of learnt GP \color{black}
        \State $A_* \leftarrow [];\ r^{sim}_* \leftarrow 0$
        \For{$k = 1\ldots \mathcal{K}$}
            \State $r^{sim}_k \leftarrow$ \textsc{PINN\_MCTS}$(a_{0:N},\ x_g,\ A_k = [],\ \mu,\ \sigma)$
            \If {$r^{sim}_k < r^{sim}_*$}
                \State $r^{sim}_* \leftarrow r^{sim}_k;\ A_* \leftarrow A_k$
            \EndIf
        \EndFor
        \State \Return $r^{sim}_*,\ A_*$
    \EndProcedure\\
    
    \Procedure{PINN\_MCTS}{$a_{i:N},\ x_g,\ A,\ \mu,\ \sigma$}
        \State \color{teal} // $a_{i:N}$ :- sub-action sequence from task decomposition \color{black}
        \State \color{teal} // $A$ :- partial sequence of selected sub-actions \color{black}
        \State \color{teal} // $x_g$ :- goal position; $\mu, \sigma$ :- mean, variance of learnt GP \color{black}
        \If{$(i < N)$}
            \State $(a^1_i, a^2_i, \ldots, a^\mathcal{D}_i) \leftarrow a_i$ \color{violet} \Comment{Sampled sub-actions} \color{black}
            \State $j_* \leftarrow \underset{j=1\ldots \mathcal{D}}{argmax} ~ v^j_i + \alpha \sqrt{\left(\log{\underset{\mathcal{D}}{\Sigma} n^l_i}\right)/n^j_i}$
            \color{violet} \Comment{UCB} \color{black}
            \State $A \leftarrow A \cup {a_i^{j_*}}$ \color{violet} \Comment{Update action sequence} \color{black}
            \State $r^{sim}_i \leftarrow$ \textsc{PINN\_MCTS}$(a_{i+1:N}),\ x_g,\ A,\ \mu,\ \sigma)$
            \State $v^{j_*}_i \leftarrow (v^{j_*}_i \times n^{j_*}_i + r^{sim}_i) / (n^{j_*}_i + 1)$
            \State $n^{j_*}_i \leftarrow n^{j_*}_i + 1$
            \State \Return $r^{sim}_i$
        \Else
            \State  $x^{sim}_N \leftarrow$ \color{OrangeRed} \texttt{PINN\_Rollout}$(A)$ \color{black}
            \State $r^{sim}_N \leftarrow$ \texttt{Reward}$(x^{sim}_N - x_g)$
            \State \color{blue} $\hat{r}^{sim}_N \leftarrow r^{sim}_N + \mu (A) + \beta^\frac{1}{2} \sigma (A)$ \color{black} \color{violet} \Comment{Reward Correction} \color{black}
            \State \Return $\hat{r}^{sim}_N$
        \EndIf
    \EndProcedure
    \end{algorithmic}
\end{algorithm}

Execution in the environment (\texttt{Real\_Rollout}) is slow. \texttt{PINN\_MCTS}, thus, uses fast \color{OrangeRed} \texttt{PINN\_Rollout} \color{black} to predict the reward associated with action sequence $A$. Assuming a sequence of skill networks $\{\mathcal{F}_{\theta_1}, \mathcal{F}_{\theta_2}, \ldots, \mathcal{F}_{\theta_m}\}$, PINN predicts the trajectory of the ball-like object with state $x_o$ by cascading the networks as $\mathcal{F}_{\theta_m}(\ldots\mathcal{F}_{\theta_1}(x_o,t_1),\ldots t_m)$ leading to a final state ($PINN\_Rollout$ in Figure \ref{fig:planner_diag}). Finally, the reward is obtained based on the predicted distance between the ball-like object and the goal. The PINN-predicted trajectories are approximate and semantically evaluate the quality of an action without involving the simulator.

The lines coloured in Algorithm \ref{alg:mcts} are where we differ from the conventional MCTS. Note that, $n^j_i$ and $v^j_i$ are the number of trials and the value of sub-action $a^j_i$, respectively. The planner expands the Monte Carlo Tree for $\mathcal{K}$ iterations with typical $\mathcal{K} = 10$. Then, it executes the best action reasoned using \texttt{PINN\_Rollout} in the environment via \texttt{Real\_Rollout} to get a reward $r^{real}$. To compensate for the systematic errors, the planner finetunes the reward predictor to \color{blue}  \emph{adapt} \color{black} to the world reward model by approximating the error as a Gaussian Process ($\mathcal{GP}$). Given $t-1$ action-discrepancy pairs, $\mathcal{GP}$ provides a Gaussian posterior $\mathcal{N}(\mu_{t-1}, \sigma_{t-1})$ over the discrepancy for the next action $A_t \in \mathcal{A}$ \cite{moses2020visual}. \texttt{PINN\_MCTS} uses $\mathcal{GP}$ to correct the reward predictor in future trials
using the equation 
\begin{equation}
    r^{sim}_t = r^{sim}_t + \mu_{t-1}(a_t) + \beta^{1/2}\sigma_{t-1}(a_t)
\end{equation}
where $\mu_{t-1}(A_t)$ and $\sigma_{t-1}(A_t)$ are the mean and variance of the function value at $A_t$ at time $t-1$, and $\beta$ is a hyperparameter; we use $\beta = 0.25$. The \color{blue} \emph{reason-adapt} \color{black} cycle repeats for a few trials till the goal is achieved. Empirically, we find $\leq 5$ trials are sufficient.

\section{Evaluation Setup}
\label{sec:evaluation-setup}
\subsection{Simulation Environment and Training Details}
%
We encode four physical reasoning tasks: \emph{Launch}, \emph{Slide}, \emph{Bounce}, and \emph{Bridge} (Figure \ref{fig:skill-chaining}) inspired from ~\cite{allen2020rapid,bakhtin2019phyre} in the Isaac Gym ~\cite{nvidia2022isaacgym} environment with a simulated Franka Emika Arm interacting with objects. A simulated RGB camera captures a video of the robot performing task rollouts. The training data points are extracted at various time steps in each rollout. The state of the interacting objects is captured by performing instance segmentation on the RGB image using frozen visual-language models (\cite{kirillov2023segment}, and ~\cite{liu2023grounding}). Segmentation of very thin objects (such as the swing of the pendulum) is carried out using edge detection performed via classical Hough-transform. Using a known camera calibration, the segmented masks are projected onto the 3D point cloud arising from the aligned depth image resulting in a estimate of the instantaneous pose and velocity (via numerical differentiation) of objects.
Then, each skill network is realised with 8 fully connected hidden layers with 40 neurons per layer. As is standard, the ratio of training points for the data loss and the physics-based loss (collocation points) is taken as 1:4.
Training skill network on an average requires $311.27 \pm 30.37$ seconds on NVIDIA GeForce GTX 1080 gpu.
\begin{figure}[tb]
    \centering
        \includegraphics[width=0.5\linewidth]{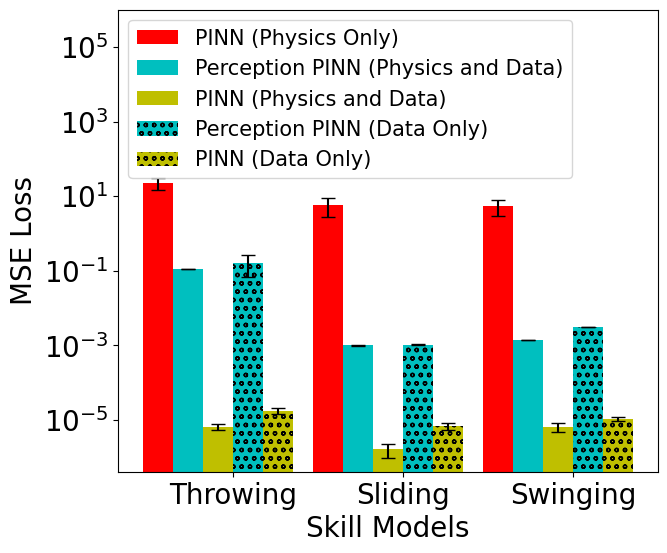}
        \includegraphics[width=0.48\linewidth, height=0.425\linewidth]{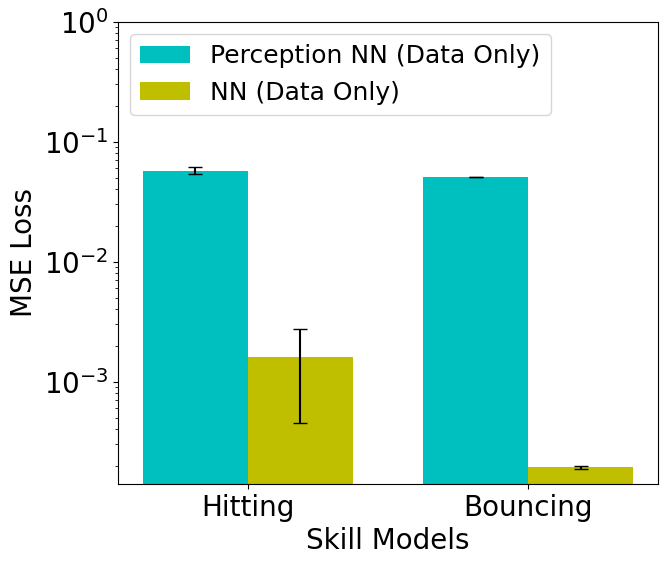}
        \caption{\footnotesize \textbf{Predictive Accuracy of Skill Networks.} Comparing prediction loss on validation dataset. PINN (Physics and Data) performs better than other networks due to guiding physics and exact data. Perception models are comparatively less accurate due to inherent noise in perceived data.}
        \label{fig:skill-model}
\end{figure}
\begin{figure}[tb]
    \centering
    \subfloat[Throwing Skill]{\includegraphics[width=0.305\linewidth]{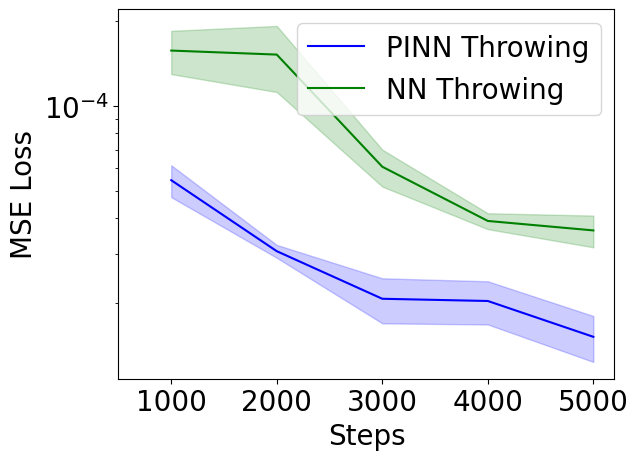}}
    \subfloat[Sliding Skill]{\includegraphics[width=0.34\linewidth]{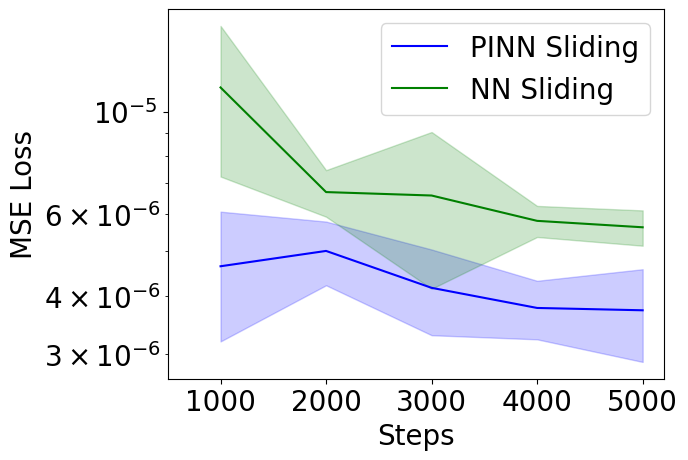}}
    \subfloat[Swinging Skill]{\includegraphics[width=0.34\linewidth]{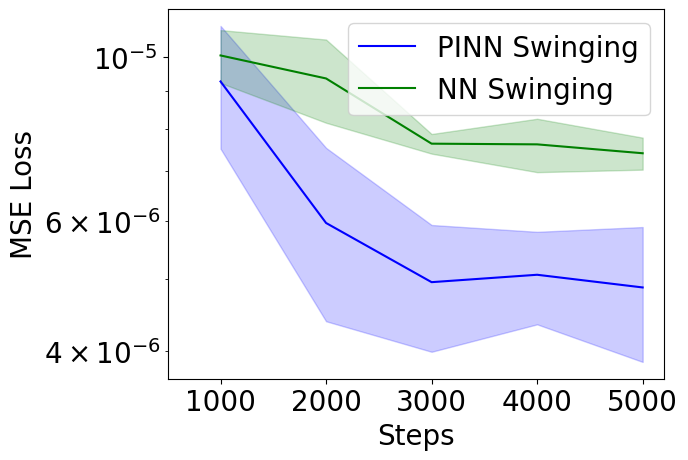}}
    \caption{\footnotesize \textbf{Training Data Efficiency.} PINN requires less training data, collected at various time steps in each rollout, to acheive the same validation loss as NN.}
    \label{fig:data-eff}
\end{figure}

The evaluation setup for task planning consists of simulated Franka Emika Robot Manipulator(s) positioned in an environment with interactable objects (within the kinematic range of at least one robot) such as a wedge, a bridge, a pendulum etc. that can be aligned, re-positioned or released. The goal region consists of a randomly placed box that cannot be moved by the robot. Typically, the box is positioned at a distance from the initial ball state such that several sequential dynamic interactions are needed for the ball to reach the goal region. Once the ball comes to halt, the initial configuration is restored (completing a trial). 
\begin{figure*}[tb]
\centering
\def\imgwidth{0.245\linewidth}
\includegraphics[width=\imgwidth]{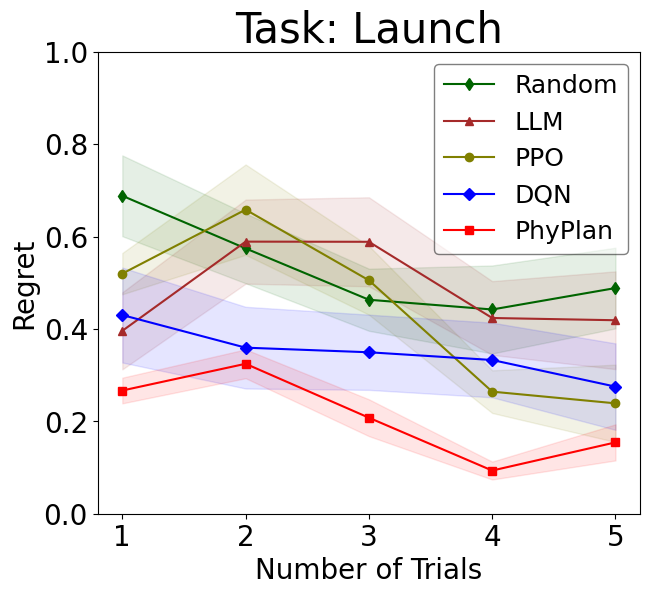}
\includegraphics[width=\imgwidth]{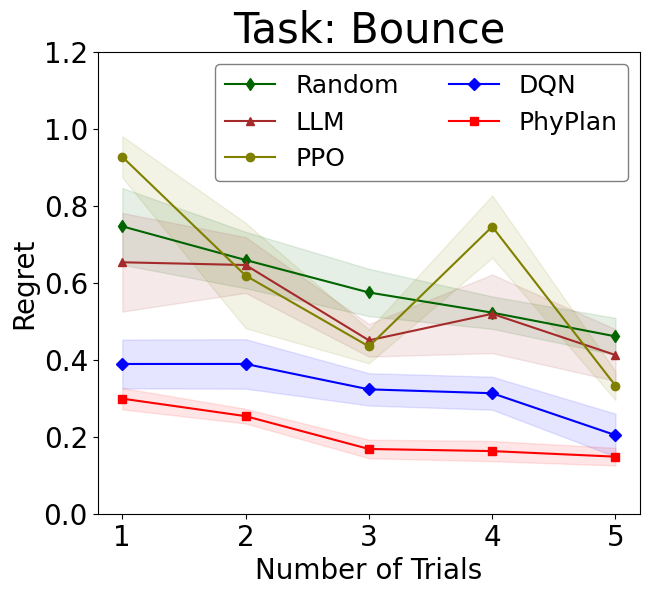}
\includegraphics[width=\imgwidth]{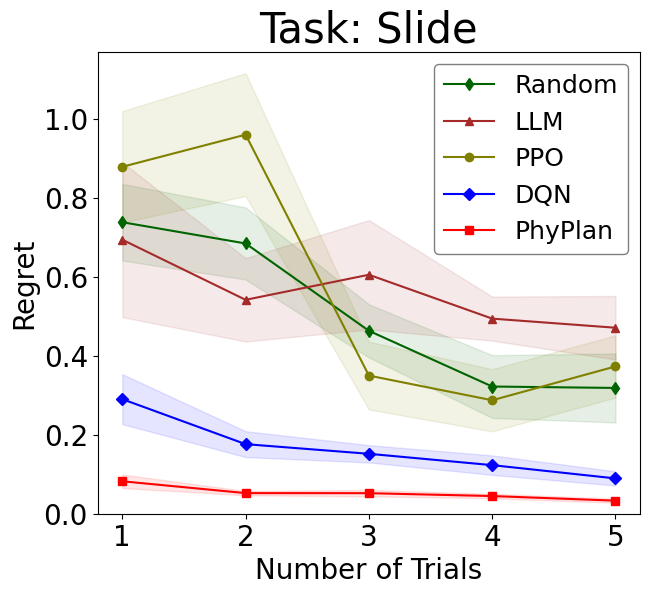}
\includegraphics[width=\imgwidth]{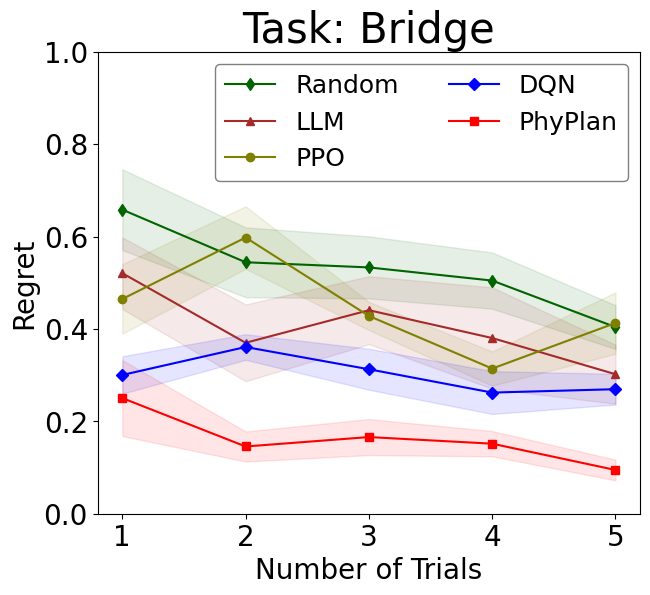}
\caption{\footnotesize \textbf{Goal-reachability for Physical Reasoning Tasks.} Comparison of minimum regret scores achieved by PhyPlan and policy learning baselines as the number of trials increases for each task. Lower is better. Each task requires compositional skill use and is \emph{a-priori} unknown to the robot. \emph{PhyPlan} consistently outperforms the baselines for all tasks. PPO, on the other hand, performs poorly, attributed to its poor training stability in setups like ours \cite{bakhtin2019phyre}.}
\label{fig:regret-results}
\end{figure*}
\subsection{Baseline Approaches for Comparison} 
Our approach, PhyPlan, contains three key components: (i) a physics-informed skill network, (ii) a reward-guided MCTS search that uses learnt skills for simulating outcomes and (iii) a GP-based adaptation approach that learns and adapts to the discrepancy between the rewards anticipated by the learnt skill networks and the real environment (high-fidelity) simulator. In our experiments, we compare our physics-informed skill network (termed PINN) with a \emph{physics-unaware} model (a standard neural network (NN) with equivalent capacity). 

Further, we compare the policy acquired by PhyPlan with the following alternative approaches for learning goal-reaching policies for physical reasoning tasks. \textbf{Random:} Samples an action uniformly at random. \textbf{LLM:} Given task description as prompt, a Large Language Model (specifically Gemini-Pro \cite{google2023gemini}) iteratively generates action and updates based on reward feedback. The prompting details can be found on the project website: \textcolor{blue}{\href{https://phyplan.github.io}{https://phyplan.github.io}}. \textbf{DQN:} a Deep Q-Network model and \textbf{PPO:} Proximal Policy Optimization model ~\cite{schulman2017proximal}, both trained using random contexts (images) and random action-reward pairs. The policy is extracted by selecting the highest scoring actions by sampling $1,000$ actions randomly as done in \cite{bakhtin2019phyre}. DQN and PPO also adapt using the Gaussian Process to account for the discrepancy from the world reward model. 
 

\section{Results}
\label{sec:results}
\subsection{Predictive Accuracy of Physical Skill Networks} 
%
%
We evaluate the prediction accuracy of the proposed physics-informed skill networks on validation data and the physical parameter estimation in unknown environments. Figure \ref{fig:skill-model} shows the skill networks trained with or without bounding boxes. We compare our Physics Informed Skill Networks with the standard Feed forward neural network (same architecture but with only data-driven loss, eq \ref{eq:data}) for (a) with perception (bounding box) and (b) without perception case (no bounding box). PINN without perception is the most efficient skill-learning network. Although, performance degrades with perception-based training due to inherent noise in data, PINN performs better than the baseline. The model with \emph{only physics loss} performs poorly due to noise in the system. 
Figure \ref{fig:data-eff} shows PINN models require less data than their NN counterparts. Therefore, incorporating physics governing equations with the data loss allows stronger generalisation and improved data efficiency compared to general \emph{physics-uninformed} time series models.

\subsection{Goal Reachability in Physical Reasoning Tasks}
Given a novel physical reasoning task, we evaluate the degree to which the robot can compose and adapt the learnt skills to attain the goal.
Figure \ref{fig:regret-results} compares the proposed model in relation to baseline models based on the average regret score~\cite{moses2020visual} attained across five trials for each task. A lower regret score indicates higher goal reachability. PhyPlan achieves lower regret than the baselines. Further, the performance gains are observed to be higher in complex tasks (e.g., the Bridge task) requiring compositional skill use and relatively lower in simpler tasks (e.g., Sliding) requiring shorter horizon reasoning. This can be attributed to (i) an accurate skill model providing rapid and reliable reward estimates and (ii) a model-based MCTS focusing the plan search towards probable plans. Further, we observe that policy-based learner PPO performs poorly compared to the DQN approach, attributed to PPO's poor training stability in setups like ours, consistent with the observation by \cite{bakhtin2019phyre}.

\begin{table}[t]
    \centering
    \begin{tabular}{|p{1.6cm}|p{1.2cm}|p{1.2cm}|p{1.2cm}|p{1.2cm}|}
        \hline
        \multirow{2}{.3\linewidth}{Models} & \multicolumn{4}{c|}{Regret} \\
        \cline{2-5}& Launch & Bounce & Slide & Bridge \\
        \specialrule{.2em}{.1em}{.01em}
        PhyPlan & 0.09 $\pm$ 0.06 & 0.15 $\pm$ 0.07 & 0.03 $\pm$ 0.02 & 0.09 $\pm$ 0.07 \\
        \hline
        PhyPlan - No Perception & \textbf{0.08 $\pm$ 0.03} & \textbf{0.10 $\pm$ 0.06} & \textbf{0.02 $\pm$ 0.02} & \textbf{0.08 $\pm$ 0.06} \\
        \hline
    \end{tabular}
    \caption{\footnotesize \textbf{Effect of perception on goal-reachability for tasks.} Table compares the final regret values obtained by \textit{PhyPlan} and \textit{PhyPlan-No Perception} over five trials for each task. The use of detected bounding boxes (perception) leads to marginal drop in performance.}
    \label{tab:perception}
\end{table}

We also implemented a \emph{PhyPlan - No Perception} planner which uses skills trained on data directly from simulator. It does not use perception to infer goal and ball positions. Table \ref{tab:perception} compares the final regret achieved by the two planners after 5 trials. \emph{PhyPlan - No Perception} performs better than \emph{PhyPlan} because of the inherent noise in Perception-based skill networks and the errors in detecting the object positions.

\begin{figure}[t] 
\centering
    \includegraphics[width=0.51\linewidth]{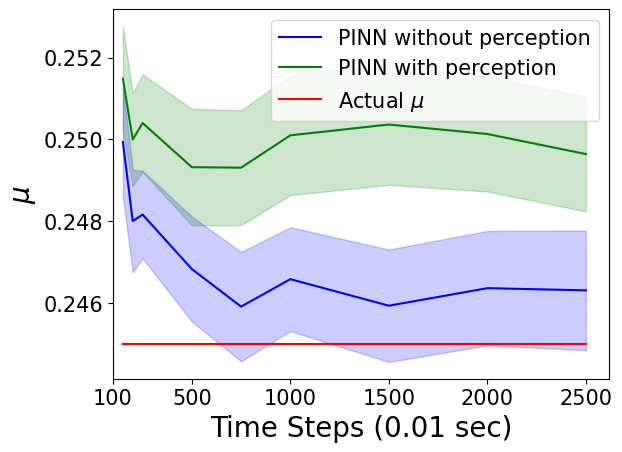}
    \includegraphics[width=0.47\linewidth]{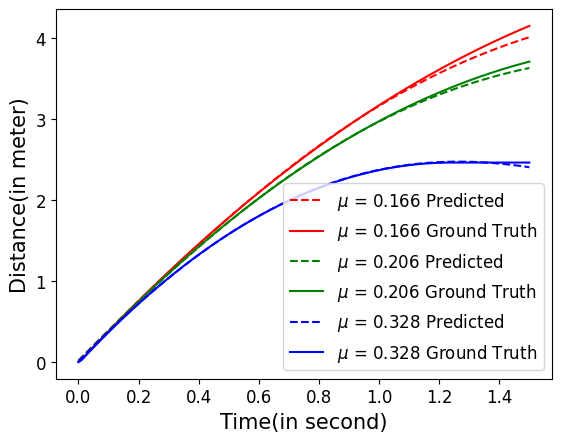}
    \caption{\footnotesize \textbf{Generalized Parameter Setting.} PINN-based approach achieves a relatively low error within a few time steps when estimating unknown $\mu$ in a single rollout in sliding skill. PINN also accurately estimates the trajectory of a box sliding on surfaces with variable $\mu$ across multiple rollouts.}
    \label{fig:unknown}
\end{figure}

\begin{figure*}[]
\centering
\def\imgwidth{0.195\linewidth}
\includegraphics[width=\imgwidth]{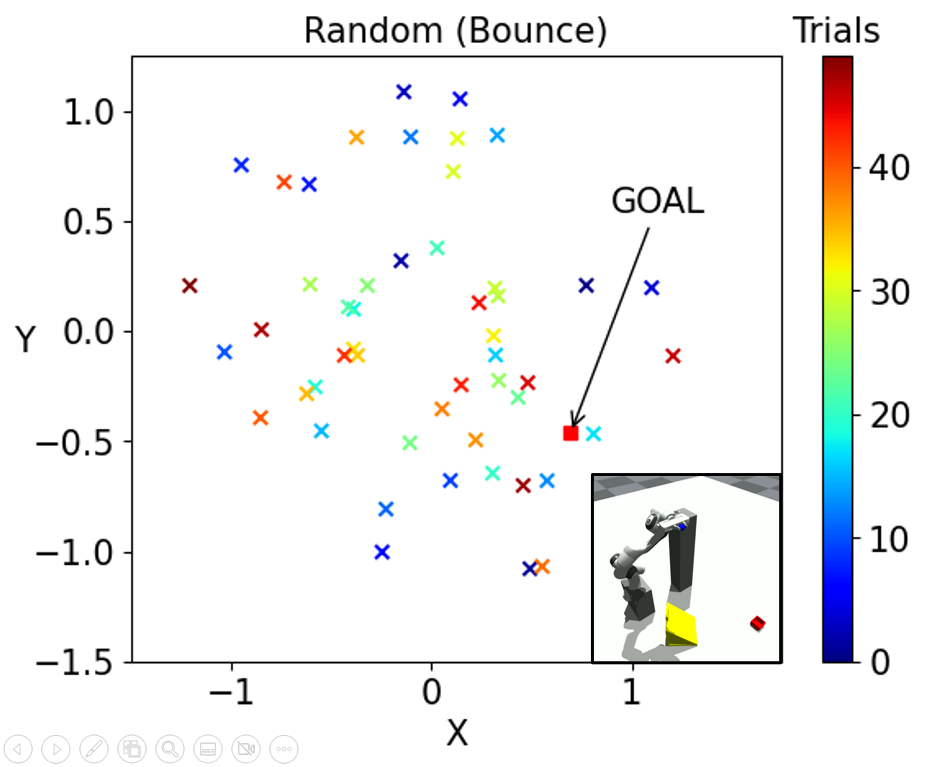}
\includegraphics[width=\imgwidth]{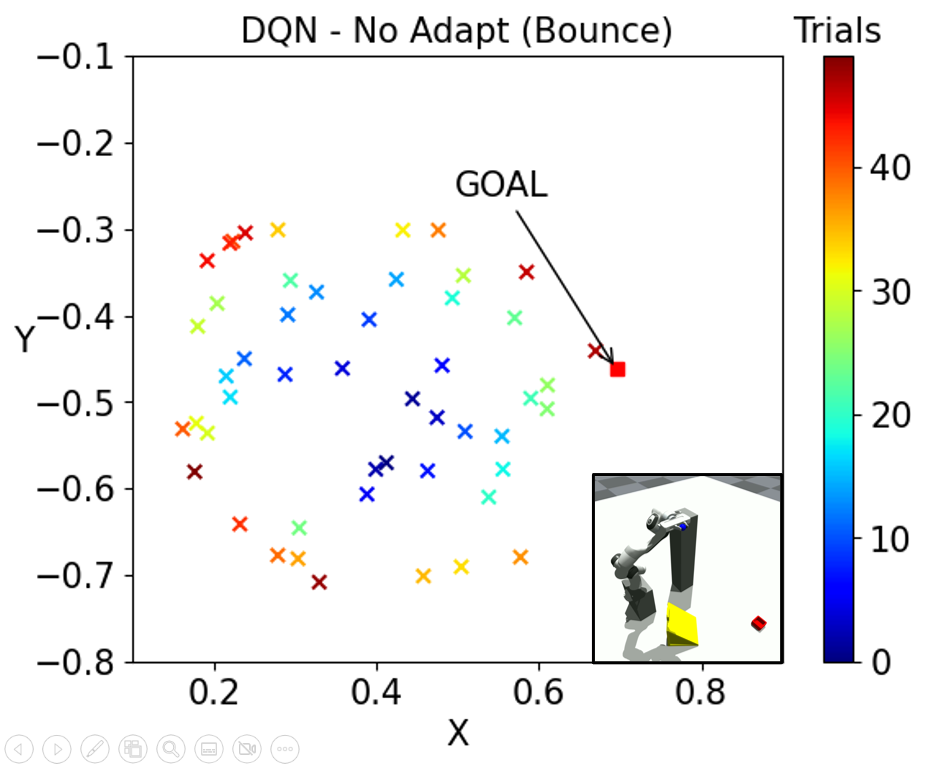}
\includegraphics[width=\imgwidth]{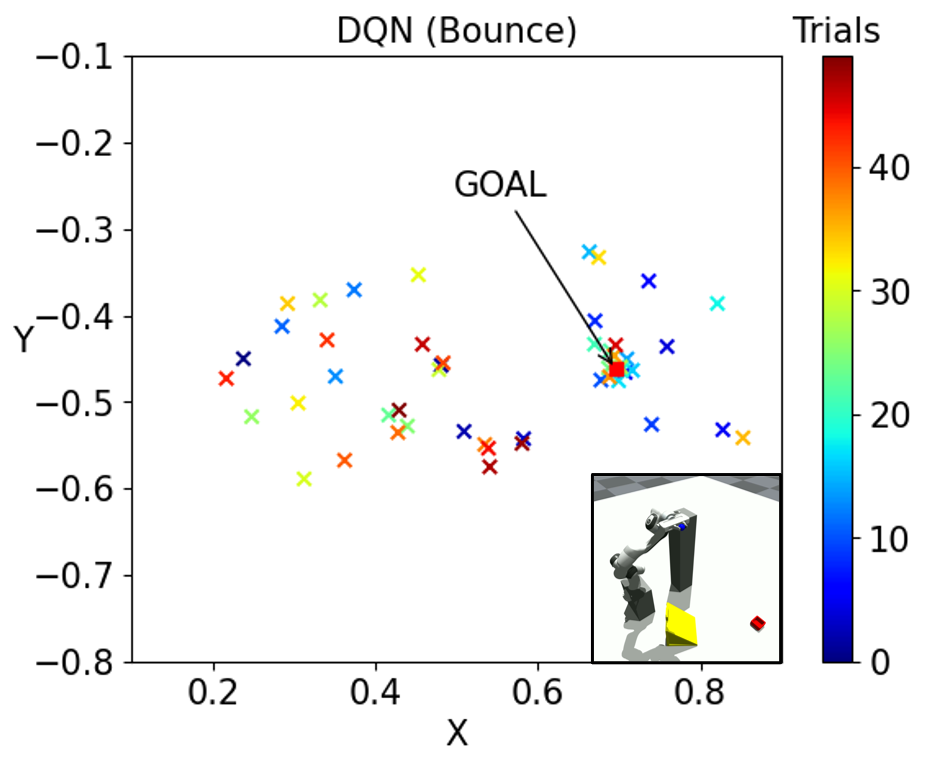}
\includegraphics[width=\imgwidth]{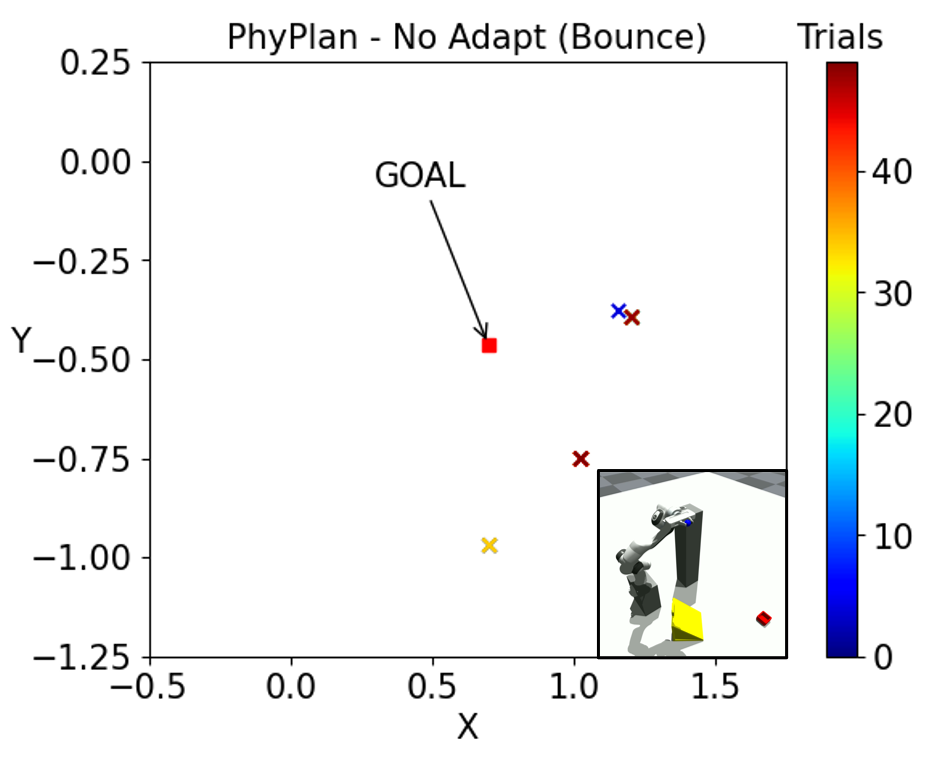}
\includegraphics[width=\imgwidth]{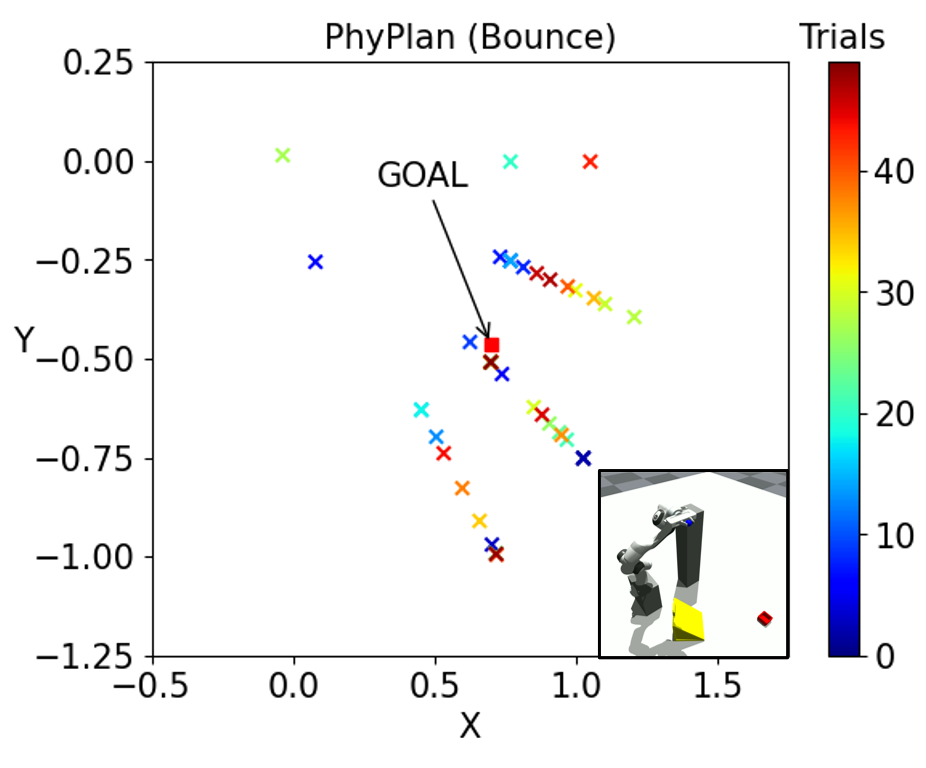}
\caption{\footnotesize \textbf{Effect of GP-based adaptation of reward model.} Showing the ball's coordinates relative to the goal in \emph{Bounce} task for 50 trials. Online Learning helps explore the action space exhaustively and converge to the optimal action by capturing the systematic errors made by the pre-trained models.}
\label{fig:online-learning}
\end{figure*}
\begin{figure*}[]
    \centering
    \includegraphics[width=0.19\linewidth, height=0.1\linewidth, frame]{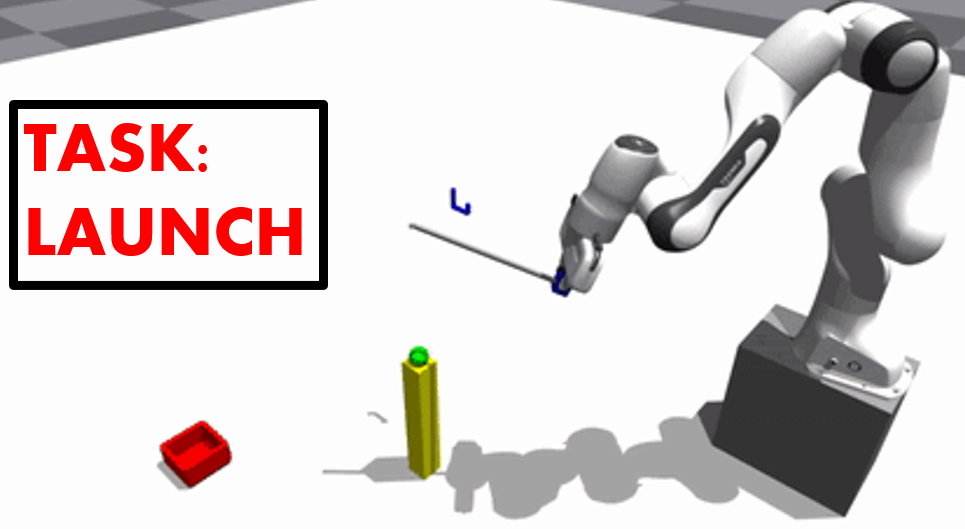}
    \includegraphics[width=0.19\linewidth, height=0.1\linewidth, frame]{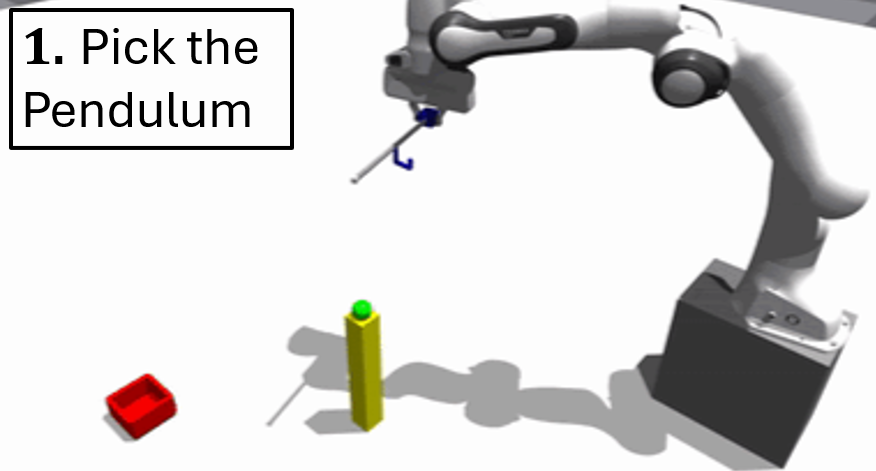}
    \includegraphics[width=0.19\linewidth, height=0.1\linewidth, frame]{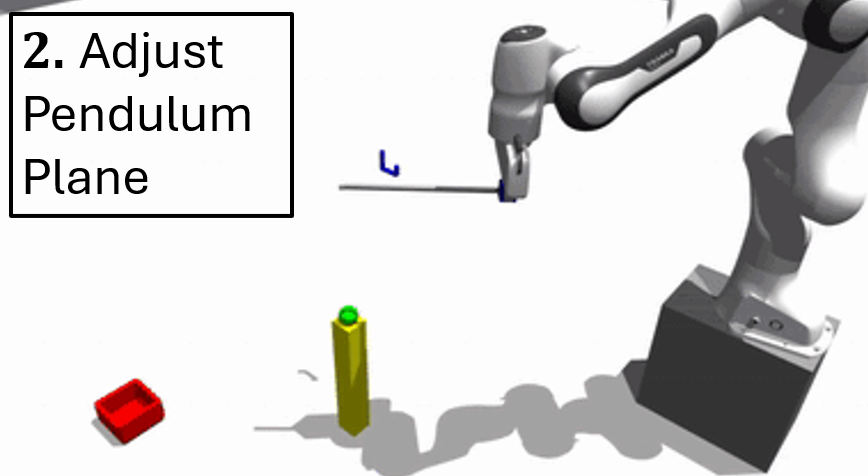}
    \includegraphics[width=0.19\linewidth, height=0.1\linewidth, frame]{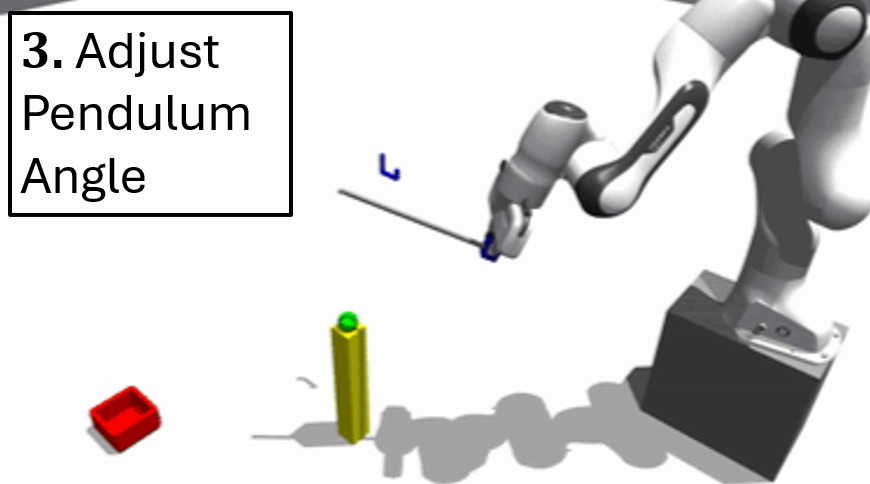}
    \includegraphics[width=0.19\linewidth, height=0.1\linewidth, frame]{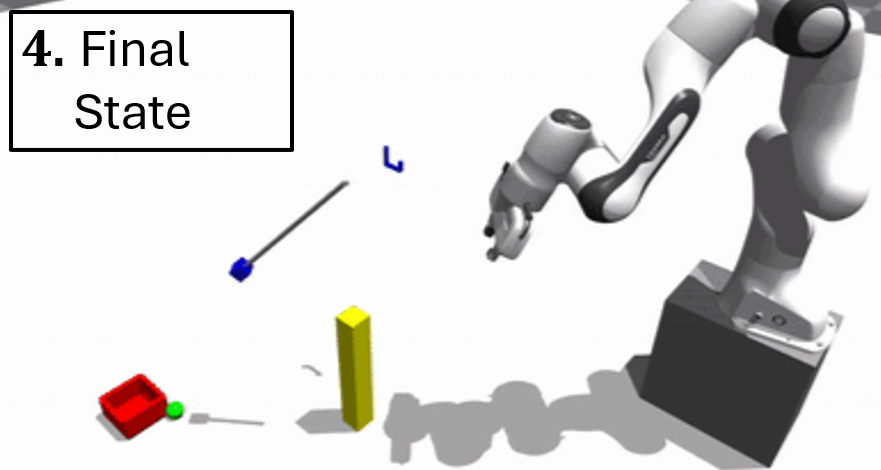}
    \caption{\footnotesize \textbf{Execution of the Policy learned for the Launch Task.} After 2 trials and 10 PINN-rollouts per trial, the robot learns a policy that correctly align the pendulum's plane and angle to throw the ball into the box.}
    \label{fig:pendulum-plan}
\end{figure*}
\begin{figure*}[]
    \centering
    \includegraphics[width=0.19\linewidth, height=0.13\linewidth, frame]{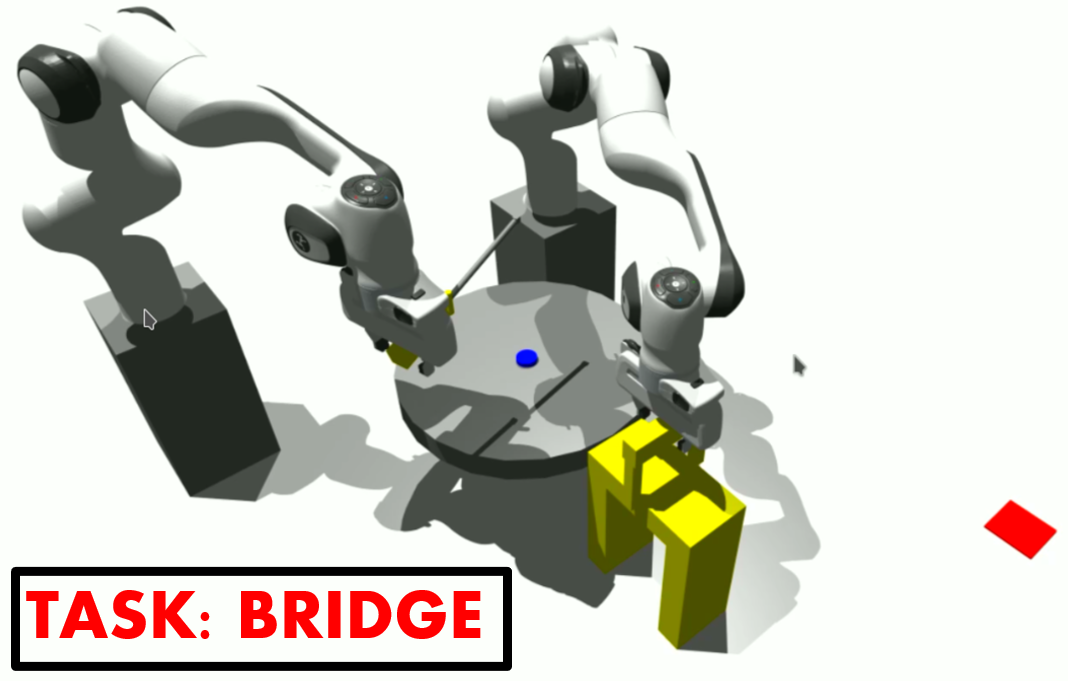}
    \includegraphics[width=0.19\linewidth, height=0.13\linewidth, frame]{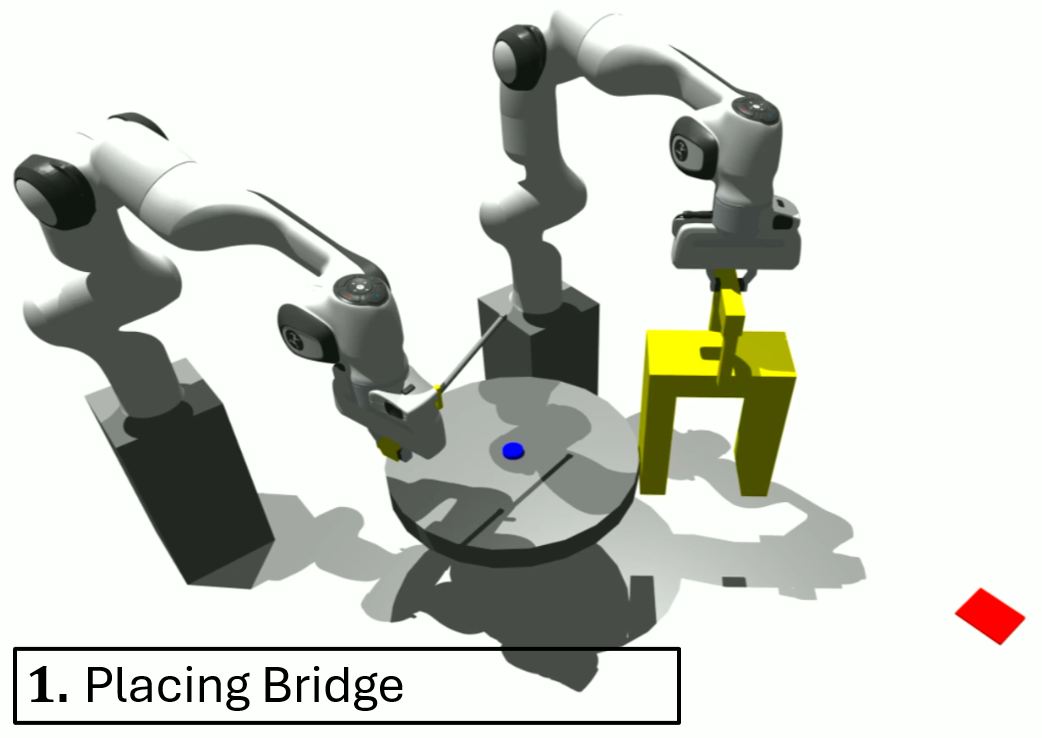}
    \includegraphics[width=0.19\linewidth, height=0.13\linewidth, frame]{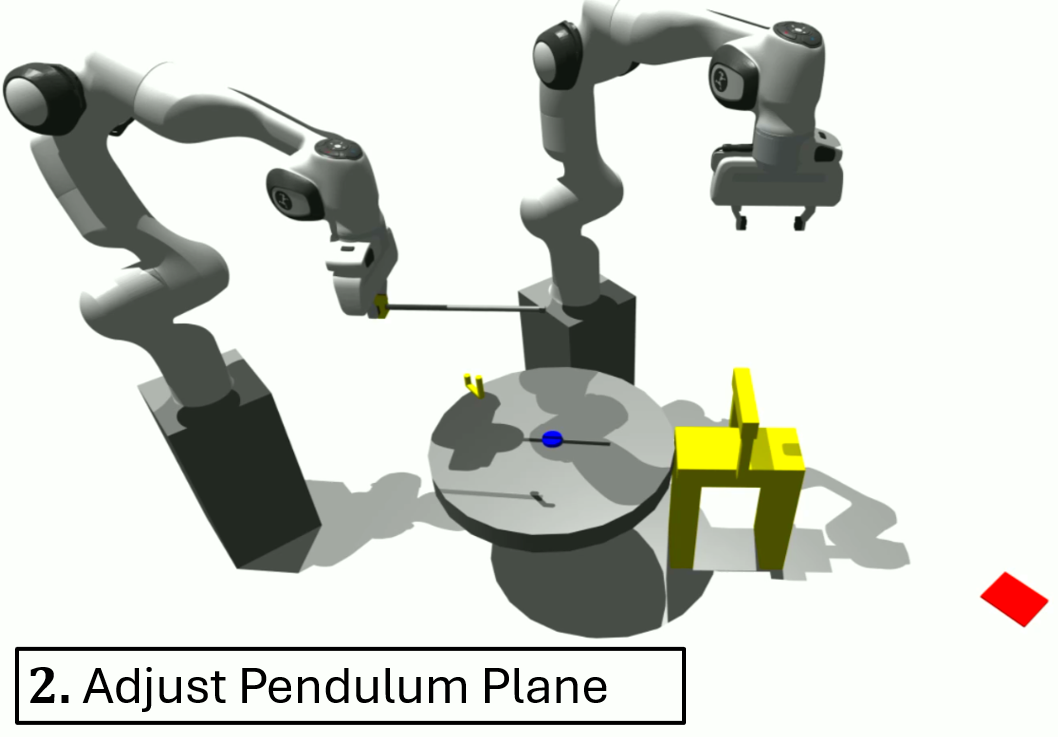}
    \includegraphics[width=0.19\linewidth, height=0.13\linewidth, frame]{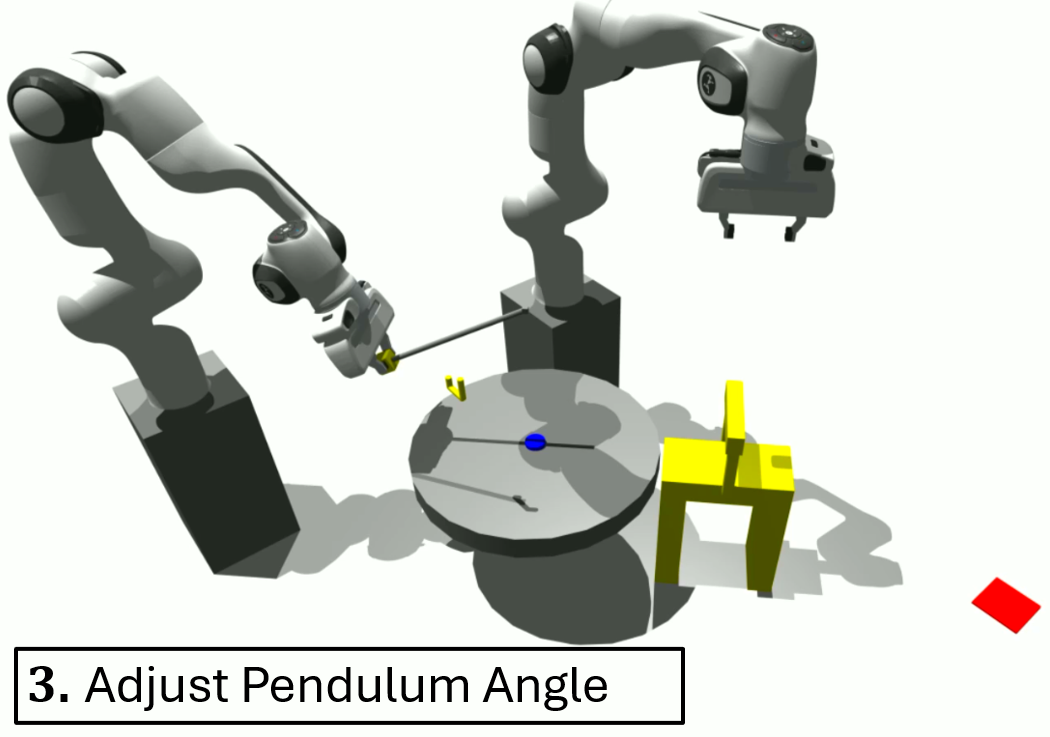}
    \includegraphics[width=0.19\linewidth, height=0.13\linewidth, frame]{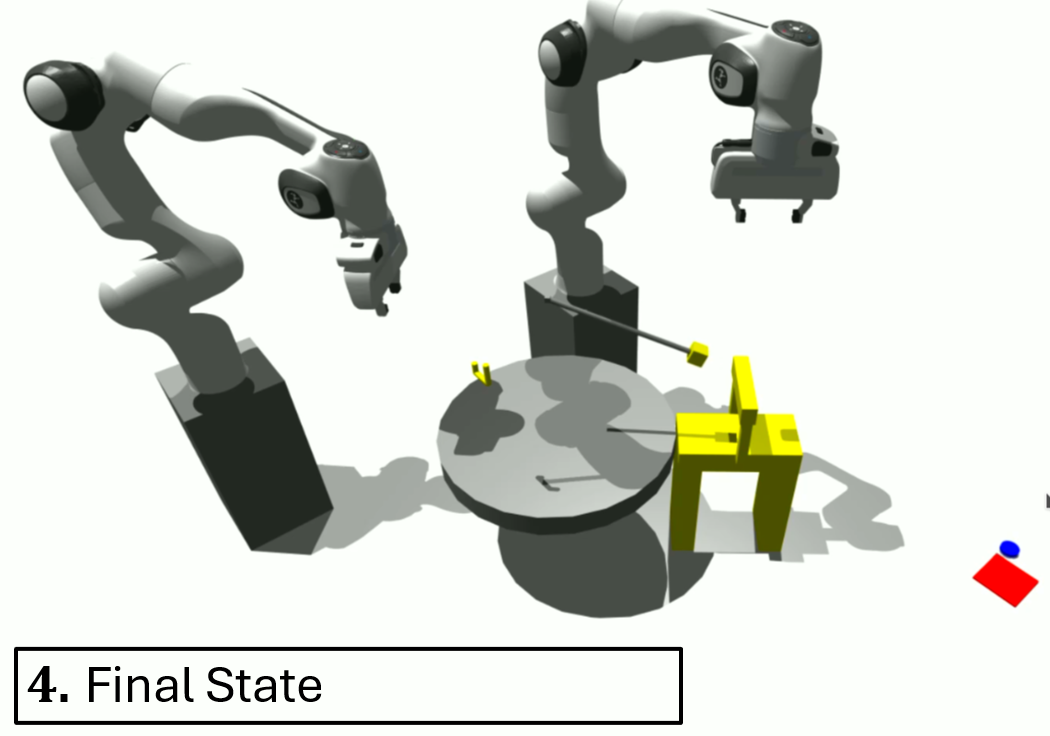}
    \caption{\footnotesize \textbf{Execution of the Policy learned for the Bridge Task.} After 5 trials and 20 PINN-rollouts per trial, the robot learns to place the bridge in the path of the box and releases the pendulum from the correct location to slide the box to the goal.}
    \label{fig:bridge-plan} 
\end{figure*}

\subsection{Generalization and Adaptation} 

%
We evaluate the degree to which the skill networks can generalize to environment without \emph{a-priori} knowledge of physical parameters during object interactions. Figure \ref{fig:unknown} (Left) indicates the the skill networks recovers the unknown physical parameters (coefficient of friction) with increasing training data points during a single roll out. The model recovers the latent parameter within a few steps in a single roll out. 
Having inferred the unknown physical parameter; as depicted in Figure \ref{fig:unknown} (Right) we can employ PINN trained on multiple rollouts with variable unknown parameters to predict trajectory, elevating its generalizability.

Further, we analyse the effectiveness of GP in adapting to the task execution environment. We create \emph{PhyPlan-No Adapt} and \emph{DQN-No Adapt} same as PhyPlan and DQN, respectively, except the GP online adaptation framework. Figure \ref{fig:online-learning} illustrates the final positions where the dynamic object (the ball) lands relative to the goal after each trial in \emph{Bounce} task. The Random model spans the whole action space, the \emph{DQN-No Adapt} is skewed as it does not learn online, \emph{DQN} converges to the goal position in later iterations, \emph{PhyPlan-No Adapt} does not explore as it does not learn the prediction error, and \emph{PhyPlan} (with GP) learns the prediction error to converge to the goal position quickly. Further, \emph{PhyPlan} learns more systematically, exploring and refining within a confined action space. 

%
\begin{table}[tb]
    \centering
    \begin{tabular}{|p{1.5cm}|p{1.3cm}|p{1.3cm}|p{1.3cm}|p{1.3cm}|}
    \hline
        \multirow{2}{.3\linewidth}{Models} & \multicolumn{4}{c|}{Time (seconds)} \\
        \cline{2-5}
         & Bridge & Launch & Slide & Bounce \\
        \specialrule{.2em}{.1em}{.01em}
        High-fidelity & 181.89 $\pm$ 102.08 & 57.13 $\pm$ 34.36 & 30.98 $\pm$ 15.46 & 115.18 $\pm$ 53.41 \\
        \hline
        Low-fidelity (PhyPlan) & \textbf{35.00 $\pm$ 0.60} & \textbf{3.80 $\pm$ 0.20} & \textbf{3.07 $\pm$ 0.15} & \textbf{7.75 $\pm$ 1.14} \\
        \hline
    \end{tabular}
    \caption{\footnotesize \textbf{Runtime Comparison.} PhyPlan's runtime is manifold less than the planner without Physics-Informed skill networks that uses a high-fidelity simulator for every rollout.}
    \label{tab:time-results}
\end{table}

\begin{figure}[]
    \centering
    \includegraphics[width=\linewidth]{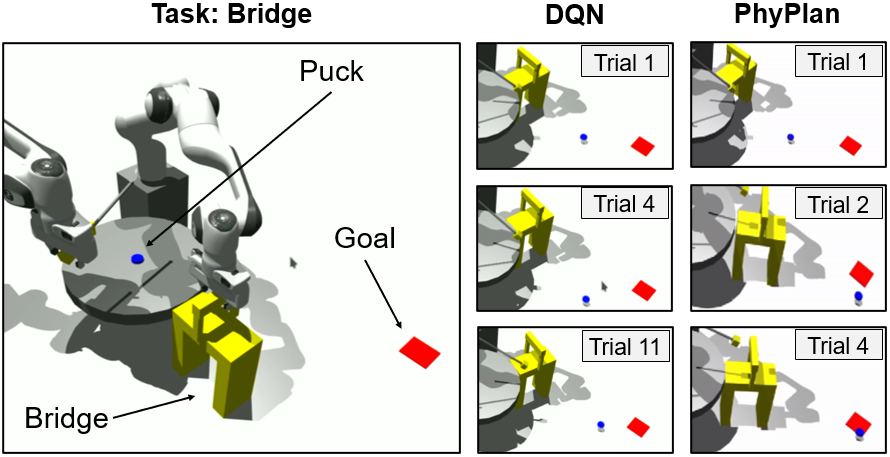}
    \caption{\footnotesize \textbf{Comparing PhyPlan with DQN.} The trials performed by DQN vs those performed by PhyPlan in \emph{Bridge} task. DQN does not use bridge even after 11 trials unlike PhyPlan which uses it starting from the second trial and perfectly aligns it in the fourth trial to achieve the goal.}
    \label{fig:dqn-comp}
\end{figure}

Table \ref{tab:time-results} compares the runtime required for PhyPlan to plan one trial of tasks with and without the use of fast simulations via learnt skills. Note that PhyPlan carefully balances using the fast learnt model with few rollouts in the real environment (the high fidelity simulator) during planning. The following table compares this approach with planning where the robot constructs the MCTS tree by \emph{actually} simulating the skills in the high-fidelity simulator (which is slow). The results indicate that use of the learnt skill networks during plan search (with occasional calls to the real simulator) significantly speeds up planning. Further, note that the simulating physical interactions such as projectile motion, sliding etc. require high computational requirements and hence the gain from learnt simulators are significant in this domain.

\subsection{Qualitative Analysis} 

Figures \ref{fig:pendulum-plan} and \ref{fig:bridge-plan} demonstrate a simulated Franka Emika robot learning to perform the tasks. The robot interacts with the objects using crane grasping with a task space controller for trajectory tracking. The figures show the robot learning (i) the pendulum's plane and release angle in the \emph{Launch} and the \emph{Bridge} tasks; and (ii) additionally the bridge's pose in the \emph{Bridge} task for the ball or puck to fall into the goal region.
The policy is learnt jointly for all objects, and the actions are delegated to the robot with the desired object in its kinematic range. Once the robots position the tools, the ball or the pendulum is dropped. The results indicate the ability to generalise to new goal locations by using objects in the environments to aid task success.

Figure \ref{fig:dqn-comp} shows a qualitative comparison between PhyPlan and DQN in the \emph{Bridge} task. PhyPlan demonstrates the capability of long-horizon reasoning in substantially less number of trials. PhyPlan has a more structured plan than DQN due to its MCTS-based exploration. Further, note that the robot learns to use the bridge effectively; a physical reasoning task reported earlier \cite{allen2020rapid} to be challenging to learn for model-free methods.

\section{Conclusions and Future Work}
We considered the problem of training a robot to perform unseen tasks requiring composition of physical skills such as throwing, hitting, sliding, etc. We leverage physics-governing equations to efficiently learn neural predictive models for skills. For goal-directed task reasoning, we embed learnt skill networks into an MCTS procedure that utilises it to perform rapid rollouts during reasoning (as opposed to using a slow full-scale physics simulation engine). Finally, we leverage GP approach for online adaptations to compensate for modelling errors via selective real rollouts. 
The approach is data efficient and performs better than the physics-uninformed approaches.
Future work will attempt to (a) learn the sequential composition of skills and plan the tasks \emph{in-situ}, (b) evaluate the model on a real robot, (c) incorporate LLMs for high level reasoning in continuous domain with sparse reward, and (d) investigate skill models for complex skills with intractable physics.


\bibliographystyle{plain}
\bibliography{refer}

\onecolumn
\newpage



\end{document}